\documentclass{article}
\usepackage{ijcai24}

\usepackage{times}
\usepackage{soul}
\usepackage{url}
\usepackage[hidelinks]{hyperref}
\usepackage[utf8]{inputenc}
\usepackage[small]{caption}
\usepackage{graphicx}
\usepackage{amsmath}
\usepackage{amsthm}
\usepackage{booktabs}
\usepackage{algorithm}
\usepackage[switch]{lineno}

\usepackage[T1]{fontenc}    
\usepackage{amsfonts}       
\usepackage{nicefrac}       
\usepackage{microtype}      
\usepackage{wrapfig}
\usepackage{slashbox}

\newcommand{\citet}[1]{\citeauthor{#1} \shortcite{#1}}
\newcommand{\citep}{\cite}

\usepackage{tikz}
\usetikzlibrary{automata, positioning, arrows}
\usetikzlibrary{arrows.meta}
\usepackage{subcaption}
\usepackage{pgfplots}
\pgfplotsset{width=10cm,compat=1.9}
\usepackage{diagbox}
\usepackage{caption}

\usepackage{etoolbox}

\newcommand{\envone}{\textsc{MultiKeyNav}}
\newcommand{\envtwo}{\textsc{CartPoleVar}}
\newcommand{\envthree}{\textsc{BasicKarel}}
\newcommand{\envfour}{\textsc{Karel}}
\newcommand{\envfive}{\textsc{PointMass}}

\usepackage{algorithm}
\usepackage{algcompatible}
\usepackage[noend]{algpseudocode}

\usepackage{times}
\usepackage{hyperref}
\usepackage{multirow,color,graphicx}
\usepackage{caption}
\usepackage{xfrac,amsfonts,amsbsy}
\usepackage[titletoc]{appendix}
\usepackage{xspace}
\usepackage{savesym,verbatim}
\usepackage[titletoc]{appendix}
\usepackage{paralist}
\usepackage{subscript}
\usepackage{tikz}
\usepackage{verbatim}
\usepackage{amsthm}
\usepackage{bbm}
\usepackage{bm}

\usepackage[inline]{enumitem}
\usepackage{lipsum}
\usepackage{mathtools}
\usepackage{cuted}
\usepackage{placeins}

\newtheorem{definition}{Definition}

\usepackage{thmtools}
\usepackage{thm-restate}

\makeatletter

\newcommand{\Rmnum}[1]{\expandafter\@slowromancap\romannumeral #1@}
\makeatother

\DeclareMathOperator*{\argmax}{arg\,max}

\def \argmax {\mathop{\rm arg\,max}}

\newcommand{\norm}[1]{\left\lVert#1\right\rVert}

\allowdisplaybreaks

\title{Learning Embeddings for Sequential Tasks Using Population of Agents}

\author{
Mridul Mahajan
\and
Georgios Tzannetos\and
Goran Radanovic\And
Adish Singla\\
\affiliations
Max Planck Institute for Software Systems\\
\emails
\{mrmahaja, gtzannet, gradanovic, adishs\}@mpi-sws.org
}

\begin{document}

\maketitle

\begin{abstract}
We present an information-theoretic framework to learn fixed-dimensional embeddings for tasks in reinforcement learning. We leverage the idea that two tasks are similar if observing an agent's performance on one task reduces our uncertainty about its performance on the other. This intuition is captured by our information-theoretic criterion which uses a diverse agent population as an approximation for the space of agents to measure similarity between tasks in sequential decision-making settings. In addition to qualitative assessment, we empirically demonstrate the effectiveness of our techniques based on task embeddings by quantitative comparisons against strong baselines on two application scenarios: predicting an agent's performance on a new task by observing its performance on a small quiz of tasks, and selecting tasks with desired characteristics from a given set of options.    
\end{abstract}

\section{Introduction}
\label{sec:intro}

Embeddings are widely used to represent data points as vectors in a space that captures meaningful relations between them \citep{cv1,cv2,cv3,nlp1,nlp2,nlp3,RW6}. They could also be utilized as representations for tasks, as studied in various areas such as multi-task learning \citep{Zhang2018LearningTM}, meta-learning \citep{DBLP:conf/iccv/AchilleLTRMFSP19}, and domain-adaptation \citep{domain2vec}. In reinforcement learning (RL), task embeddings could be used to understand the shared structure in sequential decision-making problems if similar tasks are embedded in close proximity. This enables efficient, one-shot computation of task similarity, eliminating the need for time-consuming policy rollouts.

Despite the potential benefits, prior work on learning task embeddings in RL \citep{RW1,RW2,RW4,RW13,RW18,RW20,RW23,RW9,RW14,RW3,RW12,RW17,RW19} does not explicitly optimize for task similarity. This could primarily be attributed to the lack of a general framework to measure (and reason about) similarities among sequential tasks.

\begin{figure*}[t]
    \centering
    \includegraphics[page=1, width=0.6\textwidth, trim={-0.5cm 0 2cm 0}, clip]{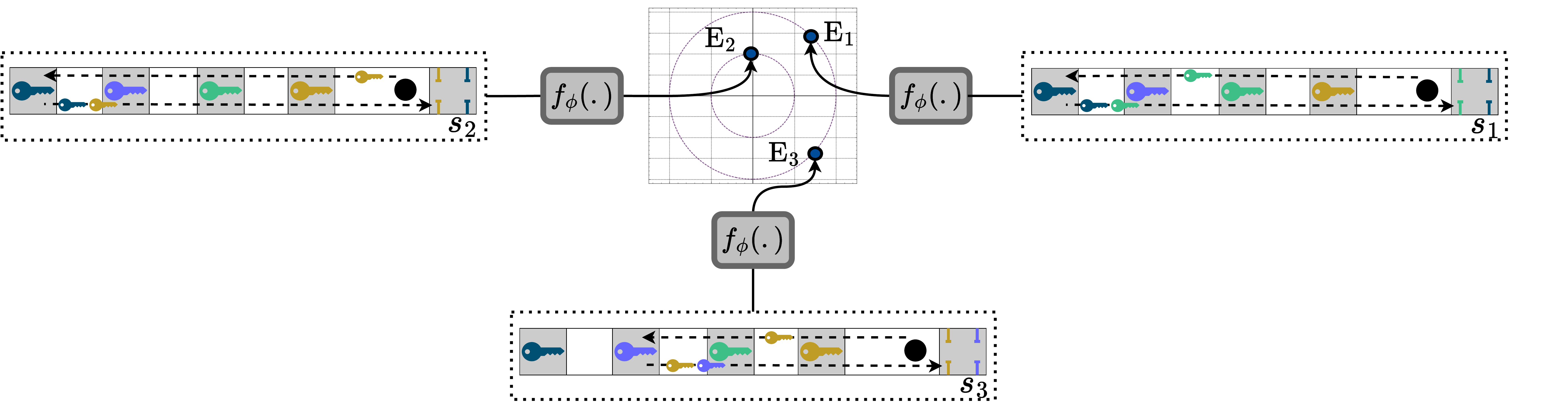}
    \caption{Schematics of our approach. We learn a task embedding function $f_{\phi}(.)$ that maps a task $s$ to its fixed-dimensional representation $\mathrm{E}$. In this illustration, we show the properties of the learned embeddings using the {\envone} environment in which tasks require the agent (shown as a black circle) to pick-up certain keys (from the gray segments) to unlock the door (the right-most segment) that has certain requirements (shown in color in the form of gates). A possible solution trajectory is depicted using dotted lines. Keys on this trajectory correspond to the ones that the agent possesses at that point in time. For instance, in task $s_{2}$, the agent starts off with the yellow key in possession already. ${\langle \mathrm{E}_{1}, \mathrm{E}_{2} \rangle}$ is greater than ${\langle \mathrm{E}_{1}, \mathrm{E}_{3} \rangle}$, since tasks $s_{1}$ and $s_{2}$ have a common requirement of picking the blue key, and thus, are similar. Additionally, $\norm{\mathrm{E}_{2}}_{2}$ is less than both $\norm{\mathrm{E}_{1}}_{2}$ and $\norm{\mathrm{E}_{3}}_{2}$, since task $s_{2}$ requires picking a single key, while tasks $s_{1}$ and $s_{3}$ require picking two keys, which makes them harder than $s_{2}$.}
    \label{fig: overview}
\end{figure*}

To this end, we introduce an information-theoretic framework to learn fixed-dimensional embeddings for tasks in RL; the inner product in the embedding space captures similarity between tasks, and the norm of the embedding induces an ordering on the tasks based on their difficulties (see Fig.~\ref{fig: overview}). A critical component of the framework is an agent population exhibiting diverse behaviors, which serves as an approximation for the space of agents. It leverages the idea that two sequential tasks are similar to each other if observing the performance of an agent from this population on one task significantly decreases our uncertainty about its performance on the other.

Essentially, there is an underlying notion of skills required to solve sequential tasks, and several tasks require some skills in common. As these skills are unknown, we measure task similarity by gathering task correlation statistics from an agent population that emerges during the multi-task policy learning process. More concretely, we introduce an information-theoretic criterion to measure task similarity (Section~\ref{infcriterion}). Through this, we construct a set of ordinal constraints on the embeddings (with each such constraint asserting the relative similarity between a triplet of tasks), and propose a training scheme for an embedding network to learn them (Section~\ref{learning_embeddings}). 

Besides assessing the learned embedding spaces through visualizations (Section~\ref{sec:expembeddings}) and comparisons with prior work (Section~\ref{sec:quant_comparison}), we ground our framework in two downstream scenarios that are inspired by real-world applications (Section~\ref{sec:downstream_apps}). Firstly, we show its utility in predicting an agent's performance on a new task given its performance on a small quiz of tasks, which is similar to assessing a student's proficiency in adaptive learning platforms via a compact quiz \citep{HeYueya2021QuizzingPU}. Several methods for curriculum design in RL rely on the agent's value function \citep{Klink2020SelfPacedDR,Theresa2021SelfPacedCRL,Yengera2021CurriculumDesign}. Thus, one concrete application of task embeddings is in black-box curriculum design, where only the agent's performance is observable, not the value function. We demonstrate that task embeddings can be used to the infer the agent's value function by observing its performance on a small quiz. Secondly, we demonstrate the application of our framework in selecting tasks with desired characteristics from a given set of options, such as choosing tasks that are slightly harder than a reference task. This is analogous to selecting desired questions from a pool for a personalized learning experience in online education systems \citep{Ahana22AdaptiveScaff}. Through comparisons with strong baselines on diverse environments, we show the efficacy of our techniques based on task embeddings.

To summarize, our work makes the following contributions:
\begin{enumerate}[label={\Roman*.},leftmargin=*]
    \item We introduce an information-theoretic framework to learn task embeddings in RL. As part of the framework, we propose a task similarity criterion which uses a diverse population of agents to measure similarity among sequential tasks (Section~\ref{infcriterion}).
    \item We propose a scheme to learn task embeddings by leveraging the ordinal constraints imposed by our similarity criterion (Section~\ref{learning_embeddings}).
    \item To assess our framework, we perform visual assessments of the learned embedding spaces, and introduce two quantitative benchmarks: \begin{enumerate*}[label=(\alph*)]
        \item agent's performance prediction, and
        \item task selection with desired characteristics (Sections~\ref{sec:expembeddings},~\ref{sec:quant_comparison},~and~\ref{sec:downstream_apps}).\footnote{GitHub repository: \url{https://github.com/machine-teaching-group/ijcai2024-task-embeddings-rl}.\label{footnote:gitcode}}
    \end{enumerate*}
\end{enumerate}

\section{Related Work}

\textbf{Task embeddings in RL.}
Several works in meta-learning and multi-task learning have explored the use of embeddings to model relationships between sequential tasks, where embeddings are either learned explicitly through objectives such as reconstruction \citep{RW4,RW13,RW20} and trajectory-based contrastive learning \citep{RW9,RW14}, or implicitly to aid generalization to new tasks \citep{RW3,RW12,RW17,RW19}. While these methods integrate embeddings with policies solely to improve performance, we propose a framework to learn general-purpose embeddings that can be used to quantify and analyze task similarities. Furthermore, in our framework, embedding computation is a one-shot operation, unlike prior work that relies on experience data from the policy for the task. These distinctions position our work as complementary to existing methods.

\textbf{Population-based techniques.}
Our framework requires a diverse agent population. This is inline with \citep{DBLP:conf/icml/FurutaMKMLNG21,pop1,pop2,pop3,RW11}, which use agent populations in the RL setting. For instance, \citet{DBLP:conf/icml/FurutaMKMLNG21} use a randomly generated agent population to empirically estimate \textit{policy information capacity}, an information-theoretic measure of task difficulty in RL.
\section{Problem Setup}
\label{sec:problem_statement}

\textbf{MDP and Tasks.}
We use the Markov Decision Process (MDP) framework to define an environment. An MDP $\mathcal{M}$ is defined as a $6$-tuple $(\mathcal{S}, \mathcal{A}, \mathcal{R}, \mathcal{T}, \mathcal{S}_{\mathrm{init}}, \gamma)$, where $\mathcal{S}$ is the state space, $\mathcal{A}$ is the action space, $\mathcal{R}:{\mathcal{S} \times \mathcal{A}} \to \mathbb{R}$ is the reward function, $\mathcal{T}:{\mathcal{S} \times \mathcal{S} \times \mathcal{A}} \to [0, 1]$ is the transition dynamics, and $ \mathcal{S}_{\mathrm{init}} \subseteq \mathcal{S}$ is the set of initial states. Each state $s \in  \mathcal{S}_{\mathrm{init}}$ corresponds to a goal-based task (for example, the goal could be to reach a specific destination in a navigation task). Our formalism of tasks as initial states is connected to the contextual MDP setting \citep{HallakContextualMDP} (see Section~\ref{sec:quant_comparison}). The reward is $0$ on all transitions but those on which a task gets completed. On task completion, the reward is $1$. As an alternative to discounting, at each time step, there is a failure probability of $1 - \gamma$, which incentivises the agent to solve the task quickly. This ensures that the cumulative reward is binary.

\textbf{Population of agents and task solvability.}
$p(\theta)$ represents a distribution over the population of agents. Concretely, it is a distribution over the agents' policy parameters. We use $\Theta$ to represent the random variable that takes on the value $\theta$. $\mathcal{O}_{s,\Theta} \in \{0,1\}$ is a Bernoulli random variable that takes on the value $1$ if, on a rollout, an agent sampled from $p(\theta)$ could successfully solve the task $s \in  \mathcal{S}_{\mathrm{init}}$ (i.e., the cumulative reward is $1$), and $0$ otherwise. We call $\mathcal{O}_{s,\Theta}$ the optimality variable for task $s$. $\textsc{PoS}(s)\coloneqq\mathop{\mathbb{E}}[\mathcal{O}_{s,\Theta}]$ denotes the probability of success on $s$, and is the complement of task difficulty.

\textbf{Task embedding space.}
Formally, we wish to learn a task embedding function (parameterized by $\phi$) $f_{\phi}: \mathcal{S}_{\mathrm{init}} \to \mathbb{R}^{n}$, for an MDP $\mathcal{M}$ and a prior over the population of agents $p(\theta)$, that maps tasks to $n$-dimensional representations. The range of $f_{\phi}(.)$ is the task embedding space.

\textbf{Objective.}
Our objective is to learn embeddings for sequential tasks with the following properties: \begin{enumerate*}[label=(\alph*)]
    \item the inner product in the embedding space captures task similarity, where the realizations of optimality variables are similar for tasks that are embedded in close proximity, and
    \item the norm of the embedding induces an ordering on the tasks based on their difficulties.
\end{enumerate*} We formalize these objectives in Section~\ref{sec:algorithms}.
\section{Learning Framework}
\label{sec:algorithms}

In Sections \ref{infcriterion}, we formally define our information-theoretic criterion to measure task similarity in RL. In Section~\ref{learning_embeddings}, we view the problem of learning task embeddings through the lens of ordinal constraint satisfaction. 

\subsection{Information-Theoretic Task Similarity Measure}
\label{infcriterion}

Our goal is to measure similarity between sequential tasks. To this end, we propose the mutual information between task optimality variables as a measure of task similarity. This metric captures the intuition that two tasks are similar to each other if observing an agent's performance on one task reduces our uncertainty about its performance on the other. We begin by formally defining performance uncertainty. Thereafter, we provide a formal definition of our task similarity criterion.

\begin{definition}[Performance Uncertainty]\label{Entropy}
The entropy of the population with prior $p(\theta)$ solving a task $s$ is defined as:
\[\mathcal{H}(\mathcal{O}_{s, \Theta}) = - \sum_{o \in \{0, 1\}}\mathrm{P}(\mathcal{O}_{s, \Theta} = o)\log{\mathrm{P}(\mathcal{O}_{s, \Theta} = o)},
\] where $\mathcal{O}_{s, \Theta}$ is the optimality variable for s.
\end{definition}

Thus, the similarity between tasks $s_{i}, s_{j} \in  \mathcal{S}_{\mathrm{init}}$ could be measured as the reduction in $\mathcal{H}(\mathcal{O}_{s_{i}, \Theta})$ by observing $\mathcal{O}_{s_{j}, \Theta}$.

\begin{definition}[Task Similarity]\label{MI}
Given a prior over the population of agents $p(\theta)$, we measure the similarity between two tasks $s_{i}, s_{j} \in  \mathcal{S}_{\mathrm{init}}$ as the mutual information $\mathcal{I}(.;.)$ between their optimality variables $\mathcal{O}_{s_{i},\Theta}$, $\mathcal{O}_{s_{j},\Theta}$: \[\mathcal{I}(\mathcal{O}_{s_{i},\Theta}; \mathcal{O}_{s_{j},\Theta}) = \mathcal{H}(\mathcal{O}_{s_{i}, \Theta}) - \mathcal{H}(\mathcal{O}_{s_{i},\Theta} \mid \mathcal{O}_{s_{j}, \Theta}).\] 
It quantifies the information obtained about $\mathcal{O}_{s_{i},\Theta}$ by observing $\mathcal{O}_{s_{j},\Theta}$.
\end{definition}

A comprehensive algorithm detailing the empirical estimation of $\mathcal{I}$ is provided in Appendix~\ref{sec:app.algo_estimate}. Given an MDP $\mathcal{M}$ and a prior distribution of the agent parameters $p(\theta)$, our algorithm uses $\mathtt{N}$ samples to estimate $\mathcal{I}(\mathcal{O}_{s_{i},\Theta}; \mathcal{O}_{s_{j},\Theta})$. For each sample, it randomly samples $\theta_{l} \sim p(\theta)$, and performs rollouts of $\pi_{\theta_{l}}$ from $s_{i}$ and $s_{j}$ to obtain estimates of the probability mass functions required for computing $\mathcal{I}$. The estimation procedure can be invoked with the signature \textsc{Estimate}{(${s}_{i}$, ${s}_{j}$, $\mathcal{M}$, $\pi$, $p(\theta)$, $\mathtt{N}$}).

\begin{algorithm}[t]
    \caption{Learn the Task Embedding Function ($f_{\phi}$)}
    \begin{algorithmic}[1]
    \small
    \Procedure{Train}{Set of tasks $\mathcal{S}_{\mathrm{init}}$, MDP $\mathcal{M}$, Policy $\pi$, Prior distribution of the agent parameters $p(\theta)$, Number of samples $\mathtt{N}$, Hyperparameter $\lambda$, Number of iterations $\mathtt{M}$}
        \State Initialize $\phi$.
        \For{$i \in \{1, \dots, \mathtt{M}\}$}
            \State Sample task $s_{1}, s_{2}, s_{3} \sim \mathcal{S}_{\mathrm{init}}$.
            \State $\mathrm{E}_{1}, \mathrm{E}_{2},  \mathrm{E}_{3} \gets f_{\phi}(s_{1}), f_{\phi}(s_{2}), f_{\phi}(s_{3})$
            \State $\hat{\mathcal{I}}_{12} \gets \textsc{Estimate}(s_{1}, s_{2}, \mathcal{M}, \pi, p(\theta), \mathtt{N})$
            \State $\hat{\mathcal{I}}_{13} \gets \textsc{Estimate}(s_{1}, s_{3}, \mathcal{M}, \pi, p(\theta), \mathtt{N})$
            \If{$\hat{\mathcal{I}}_{12} > \hat{\mathcal{I}}_{13}$}
                \State $\mathrm{loss} \gets \log({1 + \exp({{\langle \mathrm{E}_{1}, \mathrm{E}_{3} \rangle} - {\langle \mathrm{E}_{1}, \mathrm{E}_{2} \rangle}})})$
            \Else
                \State $\mathrm{loss} \gets \log({1 + \exp({{\langle \mathrm{E}_{1}, \mathrm{E}_{2} \rangle} - {\langle \mathrm{E}_{1}, \mathrm{E}_{3} \rangle}})})$
            \EndIf    
            \State Sample task $s_{4}, s_{5} \sim \mathcal{S}_{\mathrm{init}}$.
            \State $\mathrm{E}_{4}, \mathrm{E}_{5} \gets f_{\phi}(s_{4}), f_{\phi}(s_{5})$
            \If{$\textsc{PoS}(s_{4}) > \textsc{PoS}(s_{5})$}
                \State $\mathrm{loss} \gets \mathrm{loss} + \lambda\,{\log({1 + \exp({\norm{\mathrm{E}_{4}}_{2} - \norm{\mathrm{E}_{5}}_{2})}})}$
            \Else
                \State $\mathrm{loss} \gets \mathrm{loss} + \lambda\,{\log({1 + \exp({\norm{\mathrm{E}_{5}}_{2} - \norm{\mathrm{E}_{4}}_{2})}})}$
            \EndIf   
            \State Update $\phi$ to minimize $\mathrm{loss}$.
        \EndFor
        \State \Return $\phi$    
    \EndProcedure  
    \end{algorithmic}
    \label{alg:trainf}
\end{algorithm}

\subsection{Learning Task Embeddings}
\label{learning_embeddings}

With the task similarity criterion defined, we are interested in learning a task embedding function $f_{\phi}:\mathcal{S}_{\mathrm{init}}~\to~\mathbb{R}^{n}$ (consequently, an embedding space) that satisfies the desiderata introduced in Section~\ref{sec:problem_statement}. To this end, we pose the problem of learning $f_{\phi}(.)$ as an ordinal constraint satisfaction problem. Essentially, the similarity criterion $\mathcal{I}$ imposes a set $\mathcal{C}_{\mathrm{MI}}$ of triplet ordinal constraints on the task embeddings. $\textsc{PoS}(.)$ imposes another set $\mathcal{C}_{\mathrm{NORM}}$ of pairwise ordinal constraints. 

Concretely, $\mathcal{C}_{\mathrm{MI}}$ is a collection of ordered triplets of tasks s.t. for each $(s_{1}, s_{2}, s_{3}) \in \mathcal{\mathcal{C}_{\mathrm{MI}}}$, ${\mathcal{I}(\mathcal{O}_{s_{1},\Theta}; \mathcal{O}_{s_{2},\Theta}) > \mathcal{I}(\mathcal{O}_{s_{1},\Theta}; \mathcal{O}_{s_{3},\Theta})}$. Consequently, we would like to satisfy the constraint ${\langle f_{\phi}(s_{1}), f_{\phi}(s_{2}) \rangle} > {\langle f_{\phi}(s_{1}), f_{\phi}(s_{3}) \rangle}$. Likewise, $\mathcal{C}_{\mathrm{NORM}}$ is a collection of ordered tuples of tasks s.t. for each $(s_{1}, s_{2}) \in \mathcal{\mathcal{C}_{\mathrm{NORM}}}$, ${\textsc{PoS}(s_{1}) > \textsc{PoS}(s_{2})}$. Consequently, we would like to satisfy the constraint $\norm{f_{\phi}(s_{2})}_{2} > \norm{f_{\phi}(s_{1})}_{2}$ (embeddings for easier tasks have smaller norm).

We learn the task embedding function $f_{\phi}(.)$, for an MDP $\mathcal{M}$ and a prior over the agent population $p(\theta)$, by optimizing the parameters $\phi$ to maximize the log-likelihood of the ordinal constraints under the Bradley-Terry-Luce (BTL) model \citep{Luce59}. Concretely, given a task triplet $(s_{1}, s_{2}, s_{3})$, we define: \begin{equation*}
\resizebox{\columnwidth}{!}{
$\mathrm{P}\bigl((s_{1}, s_{2}, s_{3}) \in \mathcal{\mathcal{C}_{\mathrm{MI}}}\bigr) \coloneqq \frac{\exp\bigl({\langle f_{\phi}(s_{1}), f_{\phi}(s_{2}) \rangle}\bigr)}{\exp\bigl({\langle f_{\phi}(s_{1}), f_{\phi}(s_{2}) \rangle}\bigr) + \exp\bigl({\langle f_{\phi}(s_{1}), f_{\phi}(s_{3}) \rangle}\bigr)}.$
}
\end{equation*}
Similarly, given a tuple of tasks $(s_{1}, s_{2})$, we define: \begin{equation*}
\resizebox{\columnwidth}{!}{
$\mathrm{P}\bigl((s_{1}, s_{2}) \in \mathcal{\mathcal{C}_{\mathrm{NORM}}}\bigr) \coloneqq \frac{\exp\bigl(\norm{f_{\phi}(s_{2})}_{2}\bigr)}{\exp\bigl(\norm{f_{\phi}(s_{1})}_{2}\bigr) + \exp\bigl(\norm{f_{\phi}(s_{2})}_{2}\bigr)}.$
}
\end{equation*}

Hence, the task embedding function $f_{\phi}(.)$ is learned by solving the following optimization problem:
\small
\begin{align*}
    \min_{\phi}\Biggl[{\mathop{\mathbb{E}}_{(s_{1}, s_{2}, s_{3}) \sim {\mathcal{C}}_{\mathrm{MI}}}}{\log \biggl({1\!+\!\exp\bigl({{\langle \mathrm{E}_{1}, \mathrm{E}_{3} \rangle} - {\langle \mathrm{E}_{1}, \mathrm{E}_{2} \rangle}}\bigr)}\biggr)} \\ + \lambda\!\!\!\!\!{\mathop{\mathbb{E}}_{(s_{4}, s_{5}) \sim {\mathcal{C}}_{\mathrm{NORM}}}}{\log\biggl({1\!+\!\exp\bigl({\norm{\mathrm{E}_{4}}_{2} - \norm{\mathrm{E}_{5}}_{2}\bigr)}}\biggr)}\Biggr],
\end{align*}
\normalsize
where $\mathrm{E}_{i}$ denotes $f_{\phi}(s_{i})$, and $\lambda$ is a hyperparameter.
The pseudocode for the proposed algorithm to learn the task embedding function $f_{\phi}(.)$ is given in Algorithm~\ref{alg:trainf}.

\begin{figure*}[t]
\centering

\begin{subfigure}[h]{0.8\textwidth}
\centering
\resizebox{\textwidth}{!}{
\begin{tabular}{c||c|c|c|c}
Environment & Task Variability & Action & State & Number of Tasks \\
\cmidrule{0-4}
\envone & Reward Function & $7$ & $\mathbb{R} \times \{0, 1\} ^ {6}$ & Infinite \\
\envtwo & Dynamics & $2$ & $\mathbb{R} ^ {5} \times \{0, 1\} \times [200]$ & Infinite \\
\envfive & Dynamics & $\mathbb{R}^{2}$ & $\mathbb{R}^{7}$ & Infinite \\
\envfour & Reward Function + Dynamics & $52$ & $\{0, 1\} ^ {51840}$ & $73688$ \\
\envthree & Reward Function + Dynamics & $6$ & $\{0, 1\} ^ {88}$ & $24000$ \\
\bottomrule[1pt]
\end{tabular}
}
\caption{Comparison of environments' complexity}
\label{tab: env_comparison_table}
\end{subfigure}
\begin{subfigure}[h]{0.8\textwidth}
\centering
\includegraphics[width=1.03\textwidth]{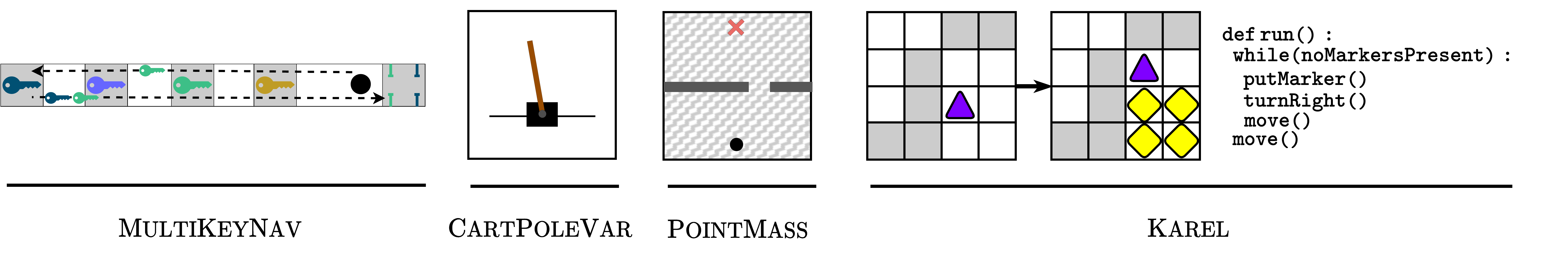}
\caption{Illustrations}
\label{fig:env_vis}
\end{subfigure}
  
\caption{We evaluate our framework on a diverse set of environments. (a) compares the characteristics of these environments. (b) illustrates these environments for a better understanding of the tasks.}
\label{fig: env_desc}
\end{figure*}
\section{Experiments: Visualizing Embedding Spaces}
\label{sec:expembeddings}

In this section, we visualize the embedding spaces to gather qualitative insights, addressing the following research questions: \begin{enumerate*}[label=(\roman*)]
\item Can distinct clusters of tasks be identified by visualizing the embedding space?
\item How does regularization through $\mathcal{C}_{\mathrm{NORM}}$ affect the embedding space?
\item What influence do agent population and environment specification have on the embedding space?
\end{enumerate*}
We begin by discussing the rationale for environment selection, describing these environments. Subsequently, we provide an overview of the embedding networks' training process, followed by the qualitative results.

\begin{figure*}[t]
\centering

   \begin{subfigure}[t]{0.36\textwidth}
 \centering
    \includegraphics[width=\textwidth,]{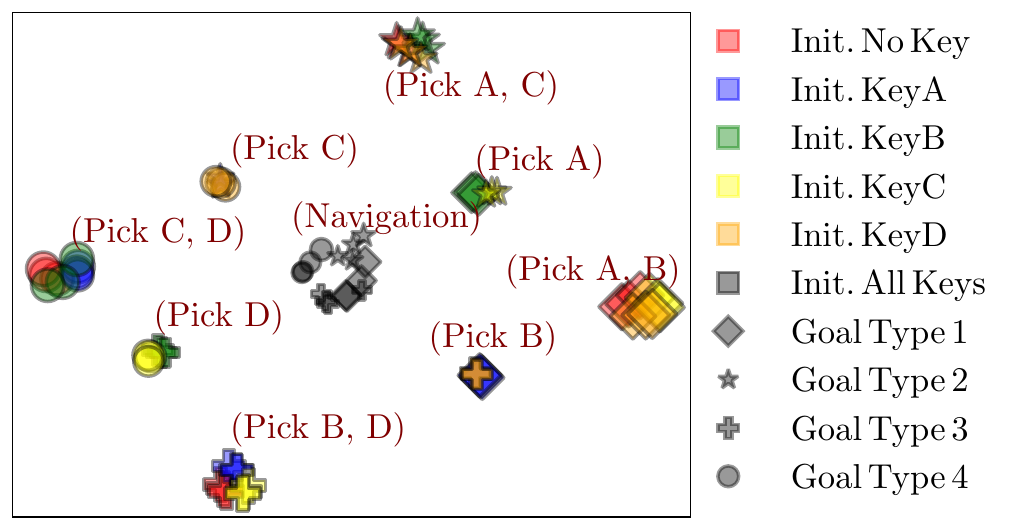}
    \caption{\envone}
    \label{fig: vis_MultiKeyNav_default}
  \end{subfigure}
 \begin{subfigure}[t]{0.275\textwidth}
  \centering
    \includegraphics[width=\textwidth]{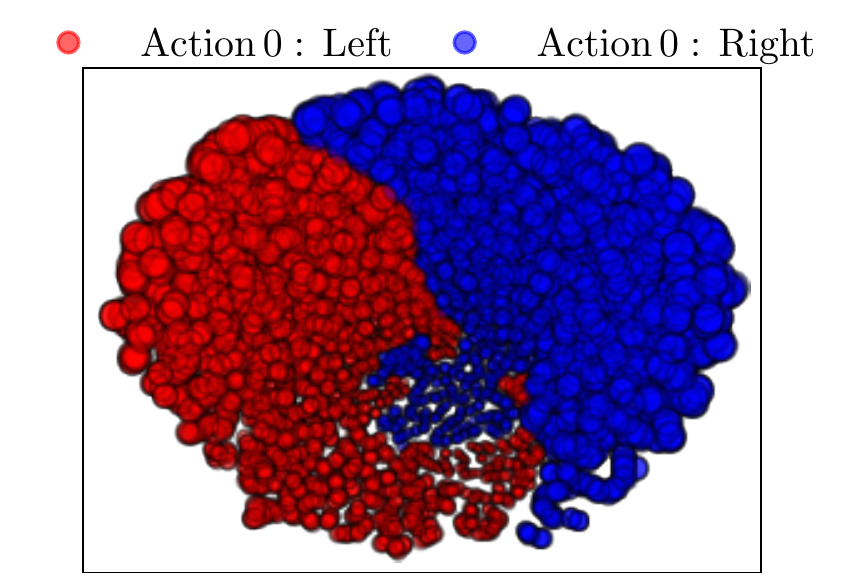}
    \caption{\envtwo}
    \label{fig: vis_CartPoleVar}
  \end{subfigure} \\
  \begin{subfigure}[t]{0.255\textwidth}
    \includegraphics[width=\textwidth]{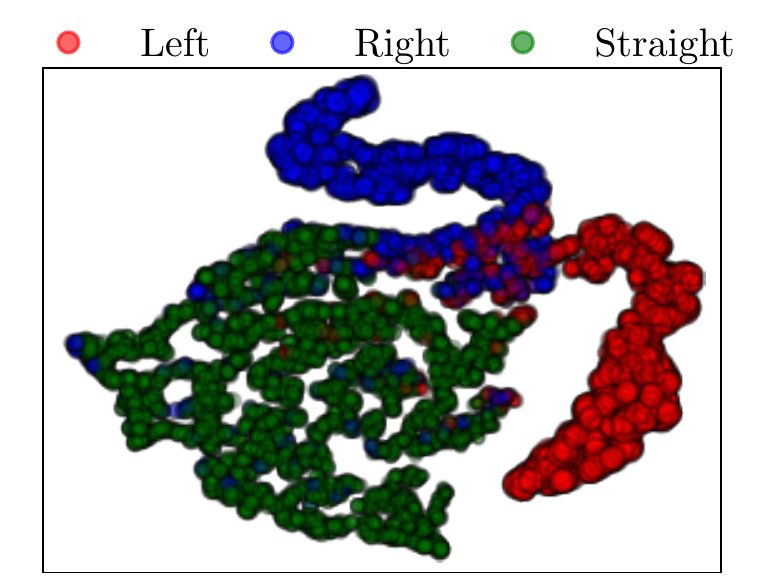}
    \caption{\envfive}
    \label{fig: vis_PointMass}
  \end{subfigure}
  \begin{subfigure}[t]{0.355\textwidth}
    \includegraphics[width=\textwidth]{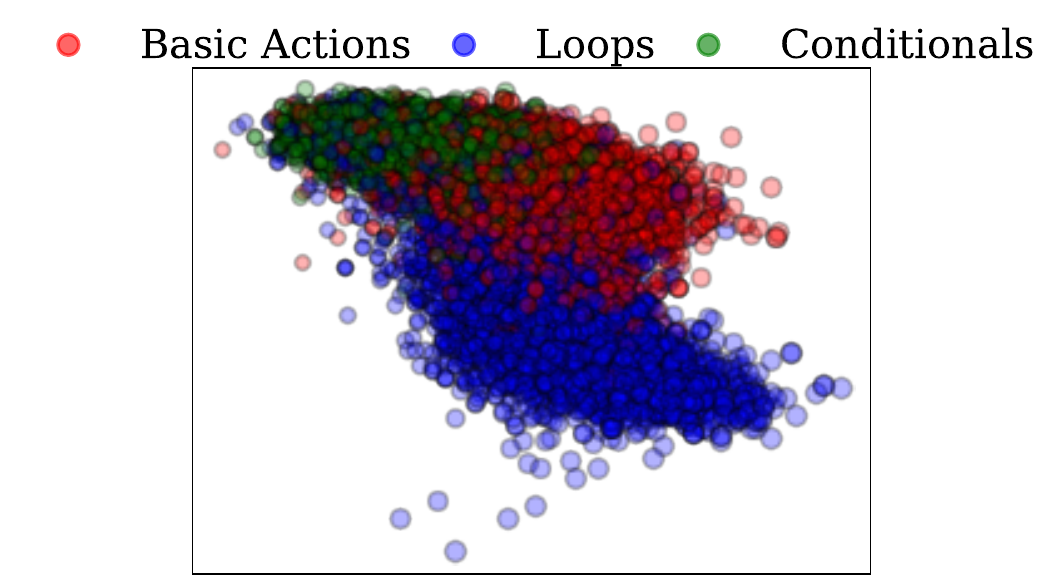}
    \caption{\envfour}
    \label{fig: vis_Karel}
  \end{subfigure}  
  \begin{subfigure}[t]{0.37\textwidth}
    \includegraphics[width=\textwidth]{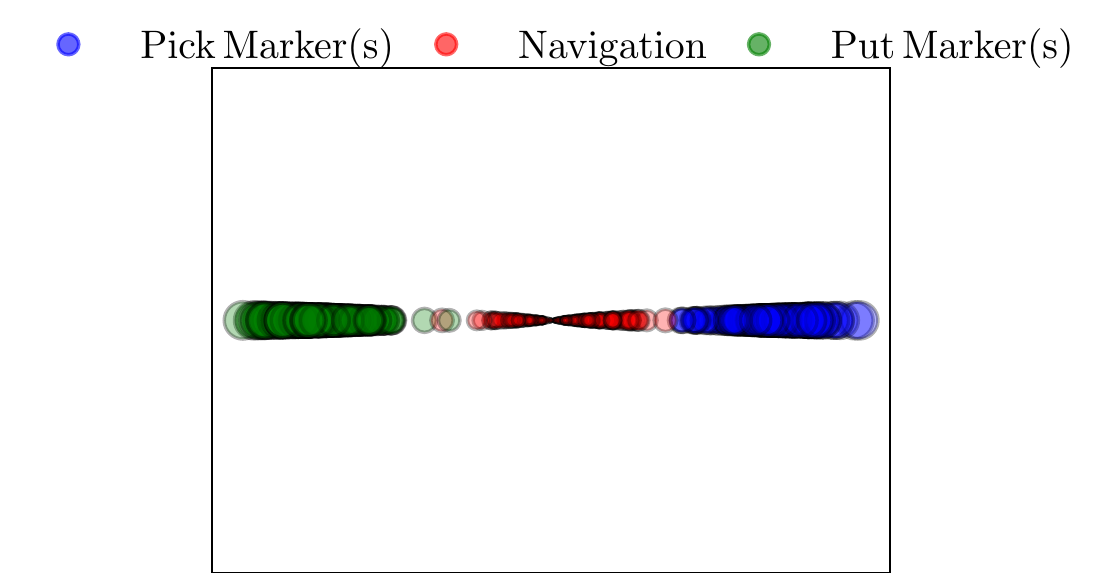}
    \caption{\envthree}
    \label{fig: vis_BasicKarel}
  \end{subfigure}
  
\caption{Visualization of the task embedding spaces learned through our framework. Each point represents a task, and the size of the points is proportional to the norm of the embeddings.}
  \label{fig: Embedding_vis_results}
\end{figure*}

\begin{figure*}[t]
\centering

\begin{subfigure}[t]{.5112\columnwidth}
 \centering
    \includegraphics[width=\textwidth,]{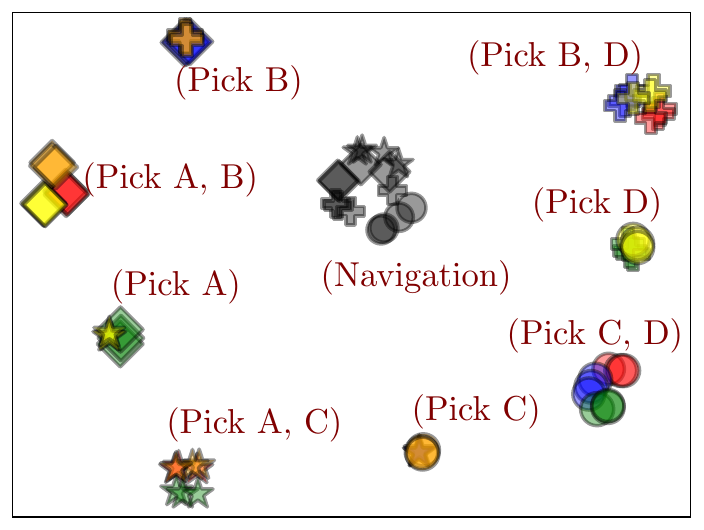}
    \caption{}
    \label{fig: vis_MultiKeyNav_default_no_norm}
  \end{subfigure}
 \begin{subfigure}[t]{.5112\columnwidth}
    \includegraphics[width=\textwidth,]{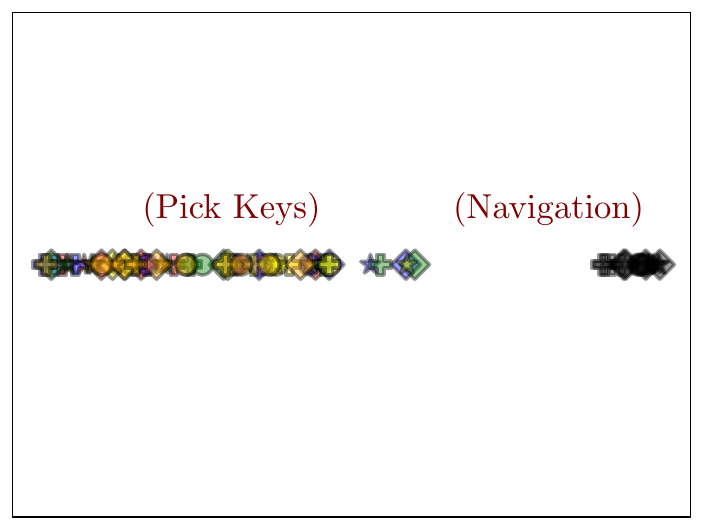}
    \caption{}
    \label{fig: vis_MultiKeyNav_maskKeys}
  \end{subfigure}  
 \begin{subfigure}[t]{.5112\columnwidth}
  \centering
    \includegraphics[width=\textwidth,]{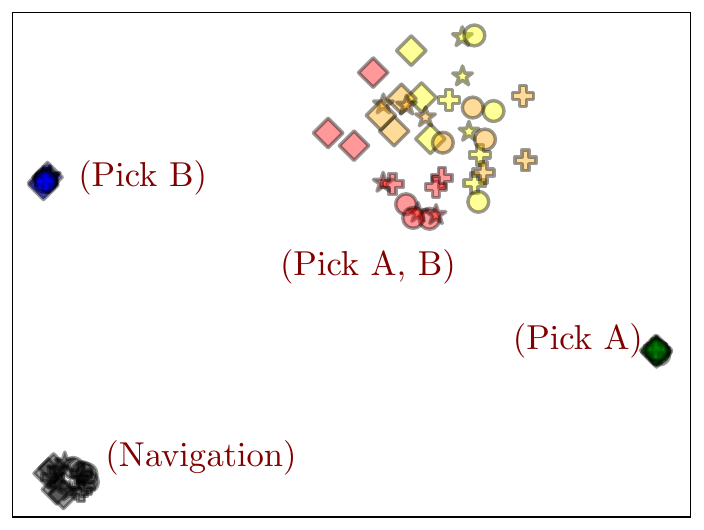}
    \caption{}
    \label{fig: vis_MultiKeyNav_AB}
  \end{subfigure}
 \begin{subfigure}[t]{.5112\columnwidth}
    \includegraphics[width=\textwidth,]{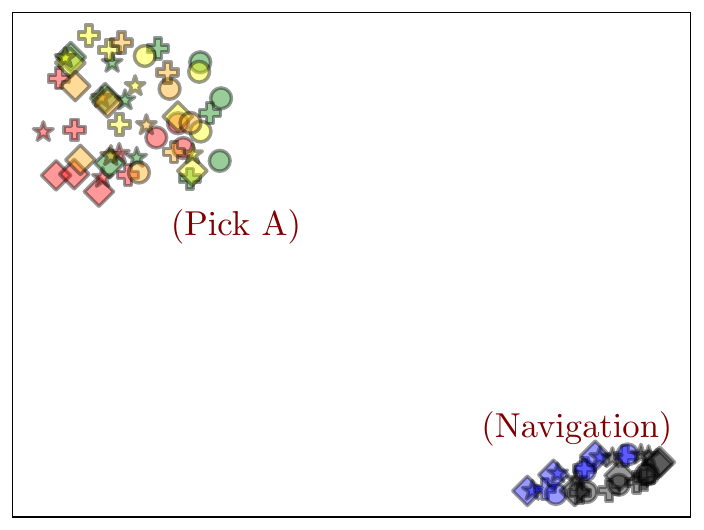}
    \caption{}
    \label{fig: vis_MultiKeyNav_A}
  \end{subfigure}
\caption{Task embedding spaces for the {\envone} environment: (a) without $\mathcal{C}_{\mathrm{NORM}}$, (b) $\mathtt{pickKey}$ actions masked, (c) all doors require $\mathtt{KeyA}$, $\mathtt{KeyB}$, and (d) all doors require $\mathtt{KeyA}$.}
  \label{fig: ablation_vis_results}
\end{figure*}
\subsection{Environments}
\label{envs}

We evaluate our framework on environments with diverse characteristics to demonstrate its generality and scalability to different sequential decision-making problems (see Fig.~\ref{fig: env_desc}). As the running example, we use {\envone} (based on \citep{devidze2021explicable}) because of its compositional nature in which the agent needs to compose different actions for picking keys (with four distinct key types, each requiring a specific action to be picked) in a task-specific manner to unlock the door. This also makes it suitable for ablation experiments. Task variability comes from the agent's initial position, the keys that it possesses initially, and the door type (with each type requiring a unique set of keys).

Given that task variability in {\envone} comes from the reward function, we use {\envtwo} to highlight our framework's applicability to environments where it comes from the dynamics instead. This environment is a variation of the classic control task from OpenAI gym \citep{gym}, and also takes inspiration from \citep{Sodhani2021MTEnv} in which the forces could be negative as well. Tasks require keeping a pole attached by an unactuated joint to a cart upright for $200$ timesteps by applying forces to the left (action $0$) or to the right (action $1$) of the cart. Task variability comes from the force $\mathtt{F}$ applied on the cart by each action, and the $\mathtt{TaskType} \in \{0, 1\}$. Tasks of Type $0$ involve ``Pulling'' with action $0$ pulling the cart from the left and action $1$ pulling the cart from the right, while tasks of Type $1$ involve ``Pushing''.

We select {\envfive} (introduced in \citep{sprl}) to test if our framework can handle continuous action spaces. Here, the agent applies forces to control a point mass inside a walled square. Tasks require reaching a fixed goal position through a gate, with variability arising from the gate width and position, along with the space's kinetic friction coefficient. 

Finally, we investigate our framework's scalability using the real-world environment {\envfour} from \citep{iclr18}, which is a challenging environment (with high-dimensional states) that has applications in programming education. Tasks require the agent to synthesize a program, potentially containing control flow constructs such as loops and conditionals, satisfying a given specification comprising input-output examples. This program serves as a controller for an avatar navigating a grid, where each cell could contain marker(s), correspond to a wall, or be empty. The avatar can traverse the grid and manipulate it by picking or placing markers. Thus, an example in the specification comprises the $\mathtt{Pre\text{-}Grid}$ and the corresponding $\mathtt{Post\text{-}Grid}$. In addition, we evaluate it on {\envthree} \citep{tzannetos2023proximal}, which is a simpler variant of {\envfour} that excludes control flow constructs.

\subsection{Training Process}
\label{training_process}

To learn the embedding function, we first obtain the agent population by taking snapshots while training a neural network policy using either behavioral cloning \citep{bc} or policy gradient methods \citep{policy_gradient_methods}. Concretely, a snapshot is recorded if the average performance on a validation set of tasks (denoted as $\mathcal{S}_{\mathrm{snap}}$) improves by $\delta_{\mathrm{snap}}$ compared to the previously recorded snapshot. A snapshot of the untrained policy is recorded by default. Different subpopulations, obtained by either masking actions or by using biased task distributions during training, are combined to form the final population. Here, masking an action corresponds to setting its logit to a large negative number. Using biased task distribution during training is another way to inject diversity into the population. In {\envone}, for instance, this could correspond to assigning low probability mass to tasks with certain types of doors in the initial state distribution during training. Finally, we parameterize the embedding function $f_{\phi}(.)$ with a neural network, optimizing its parameters as described in Algorithm~\ref{alg:trainf}. We provide additional details in Appendix~\ref{sec:app.impl_details}.


\subsection{Visualizations and Qualitative Results}
\label{visualizations}
We visualize the embedding spaces on a $2$-dimensional map using t-SNE \citep{tsne} to identify distinct clusters of tasks. Although t-SNE preserves the local structure, it does not necessarily preserve the embeddings' norm. For this reason, we scale the points in proportion to the norm of the embeddings. Additionally, we provide PCA plots in Appendix~\ref{sec:app.pca_vis}.

\textbf{Visualizations.}
For {\envone} (Fig.~\ref{fig: vis_MultiKeyNav_default}), our framework discovers distinct clusters of tasks, with each cluster corresponding to a unique set of keys that need to be picked. The norm of the embeddings is in accordance with the number of keys that need to be picked (with tasks requiring navigation only having the smallest norm). Additionally, tasks in clusters adjacent to each other share a common key requirement. For {\envtwo} (Fig.~\ref{fig: vis_CartPoleVar}), it discovers that each task exhibits one of two types of underlying dynamics. In one ($+\text{ve }\mathtt{F}$ and Type $0$, or $-\text{ve }\mathtt{F}$ and Type $1$), action $0$ moves the cart to the left, while in the other ($-\text{ve }\mathtt{F}$ and Type $0$, or $+\text{ve }\mathtt{F}$ and Type $1$), it moves the cart to the right. For {\envfive} (Fig.~\ref{fig: vis_PointMass}), it discovers three clusters of tasks based on the behavior that the agent needs to exhibit near the gate. The first cluster includes tasks in which the agent need not steer to cross the gate, while the second and third clusters contain tasks in which it must steer left or right to cross the gate, respectively. For {\envfour} and {\envthree} (Fig.~\ref{fig: vis_Karel}~and~\ref{fig: vis_BasicKarel}), it discovers different clusters based on whether the solution code requires loops or conditionals, and whether the agent needs to pick or put markers in the grid, respectively. In Appendix~\ref{sec:app.gen_exp}, we show that the learned embedding network can generalize to new tasks.

\textbf{Ablation w.r.t. $\mathcal{\mathbf{C}}_{\mathrm{\mathbf{NORM}}}$.} Fig.~\ref{fig: vis_MultiKeyNav_default_no_norm} shows the task embedding space learned without the norm ordinal constraints $\mathcal{C}_{\mathrm{NORM}}$ (i.e., $\lambda$ is set to $0$). As expected, the norm of the embeddings is not proportional to the number of keys that need to be picked. Instead, the points are nearly uniform in size.

\textbf{Ablation w.r.t. population specification.}
To understand the effect of population on the embedding space, we learn the embedding function $f_{\phi}(.)$ for {\envone} using an agent population in which $\mathtt{pickKey}$ actions are masked (Fig.~\ref{fig: vis_MultiKeyNav_maskKeys}). In this case, we obtain two distinct clusters of tasks -- one of the clusters contains tasks that cannot be solved (these require picking key(s)), and the other contains tasks that require navigation only. These results emphasize the importance of the population's quality in learning a good task embedding space.

\textbf{Ablation w.r.t. environment specification.}
We check the impact of the environment specification on the embedding space through the following {\envone} variants: \begin{enumerate*}[label=(\alph*)]
\item each door requires $\mathtt{KeyA}$ and $\mathtt{KeyB}$ (Fig.~\ref{fig: vis_MultiKeyNav_AB}), i.e., they have identical key requirements, and 
\item each door requires $\mathtt{KeyA}$ only (Fig.~\ref{fig: vis_MultiKeyNav_A}).
\end{enumerate*} Modifying the environment specification changes the task semantics, thereby impacting the embedding space. Thus, these results are inline with our intuition.


\section{Experiments: Comparison with Prior Work}
\label{sec:quant_comparison}

In this section, we compare our framework with prior work using silhouette scores, which assess clustering quality in the embedding spaces based on intuitively identified task clusters. To enable this comparison, we introduce the \textit{PredModel} baseline, inspired by prior work (e.g., PEARL \citep{RW18}). We also compare our method against embedding networks with random weights (\textit{RandomModel}).

Most existing methods use variational inference to learn latent context from task-specific experience data, where the inference network could be trained to reconstruct the MDP for the task through predictive models of reward and dynamics.  To adapt this approach to our setting, we connect our formalism of tasks as initial states to the contextual MDP setting \citep{HallakContextualMDP}, where each context (e.g., {\envone}'s context: agent's initial position, possessed keys initially, door type) corresponds to a distinct task represented by a separate MDP with context-dependent transitions and rewards. By including context variables as part of the state, this set of MDPs can be converted into an equivalent MDP, where each initial state represents a task.

The modifications needed for \textit{PredModel} are as follows: Firstly, since context is observable in our setup, we condition the approximate posterior over the embeddings on the initial state, eliminating the need for experience data. Secondly, we train the predictive models on states with context variables removed, ensuring the utilization of the embedding that the model is conditioned on. We provide additional details in Appendix~\ref{sec:app.baseline_details}.

\textbf{Results.} Table \ref{tab: silhouette_comp} reports the silhouette scores, averaged across $3$ random seeds, with $1000$ tasks per seed ($5000$ for {\envthree}). The scores for the models learned through our framework are consistently better. While \textit{PredModel} clusters similar tasks together in the embedding space for {\envtwo}, it fails to do so in rest of the environments. In contrast to {\envtwo}, where task variability comes from dense differences in the dynamics, task variability in other environments comes from sparse differences in the reward function and/or dynamics. Thus, we hypothesize that \textit{PredModel} fails on environments with sparse variability across tasks.

\begin{table}[t]
\centering
\resizebox{8.5cm}{!}{
\begin{tabular}{c||c|c|c}
Environment & RandomModel & PredModel & Ours\\
\cmidrule{0-3}
\envone & $+0.036 \pm 0.048$ & $-0.037 \pm 0.003$ & $\bm{+0.753 \pm 0.001}$ \\
\envtwo & $+0.015 \pm 0.016$ & $+0.242 \pm 0.007$ & $\bm{+0.325 \pm 0.009}$ \\
\envfive & $+0.104 \pm 0.026$ & $-0.010 \pm 0.004$ & $\bm{+0.380 \pm 0.019}$ \\
\envthree & $-0.058 \pm 0.007$ & $-0.002 \pm 0.003$ & $\bm{+0.811 \pm 0.019}$ \\
\bottomrule
\end{tabular}
} 
\caption{Comparison of silhouette scores (higher is better) based on intuitively identified clusters of tasks in the learned embedding spaces. The scores for our models are consistently better.}
\label{tab: silhouette_comp}
\end{table}

\begin{figure}[t]
\centering
    \includegraphics[width=0.8\columnwidth]{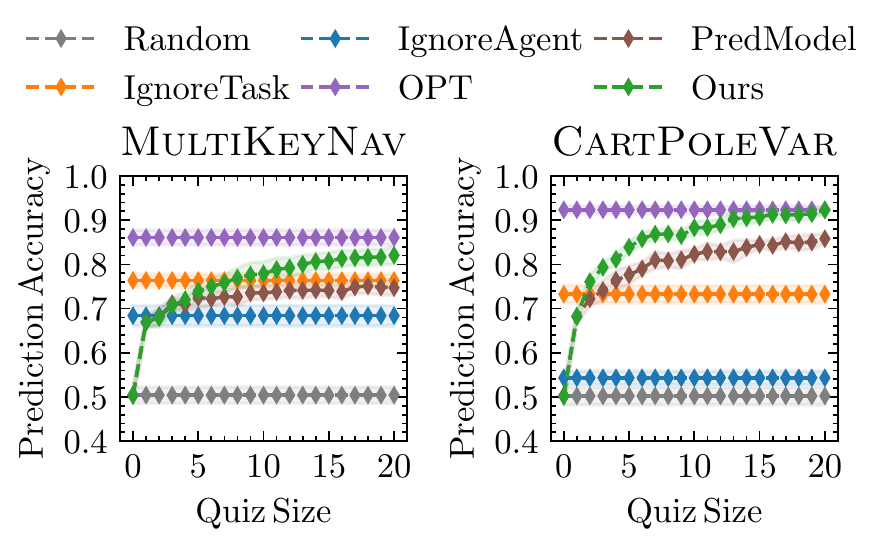}
    \caption{Results for performance prediction using task embeddings. Our method (listed as \textit{Ours}) is competitive with the \textit{OPT} baseline, which is the best one could do on this benchmark. }
    \label{fig: dt_1_results}
\end{figure}

\section{Experiments: Application Scenarios}
\label{sec:downstream_apps}
In this section, we evaluate our framework on two application scenarios: performance prediction, and task selection. We conduct this evaluation on {\envone} and {\envtwo}, as they cover two distinct sources of task variability, namely reward function and dynamics.

\begin{figure*}
    \centering
    \includegraphics[width=0.8\textwidth]{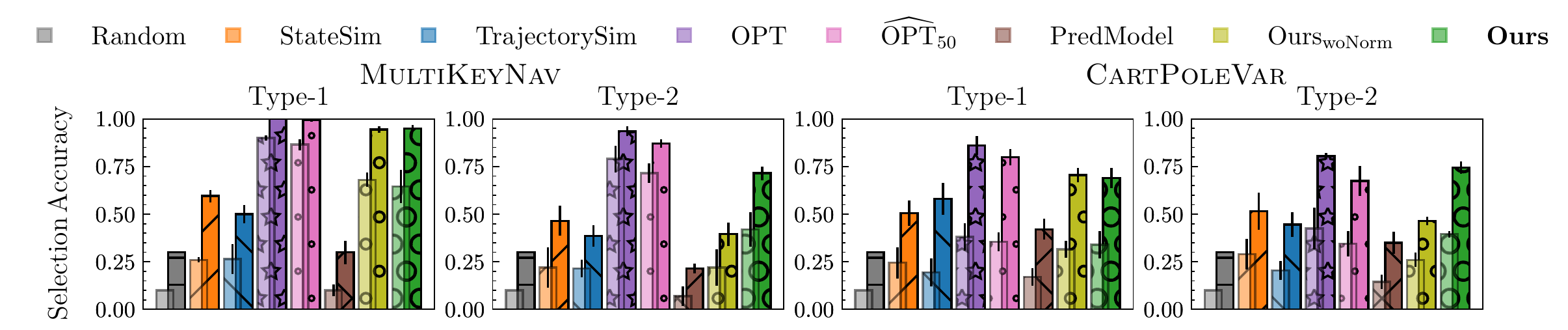}
    \caption{Task selection results using embeddings (dark bars represent Top-$3$ accuracy and light bars represent Top-$1$). Our method (listed as \textit{Ours}) is competitive with $\mathrm{\widehat{\mathit{OPT}}_\mathit{50}}$. Further, it outperforms $\textit{Ours}_{\textit{woNorm}}$ on Type-$2$ queries, highlighting $\mathcal{\mathcal{C}_{\mathrm{NORM}}}$'s significance in our framework.}
    \label{fig: dt_2_results}
\end{figure*}

\subsection{Performance Prediction}
\label{expperformance}
First, we assess the learned task embeddings by using them to predict an agent's performance on a task $s_{\mathrm{test}}~\in \mathcal{S}_{\mathrm{init}}$ after observing its performance on a quiz $\mathcal{S}_{\mathrm{quiz}}~\subseteq~\mathcal{S}_{\mathrm{init}}$. This requires estimating its probability of success on a task, which is precisely its value function when the reward is binary and sparse. Specifically, we seek to answer the following research question: Would an agent show similar performance on tasks that are in close proximity in the learned embedding space? 
We begin by creating a benchmark for this application scenario, and then compare our method against various baselines.

\textbf{Benchmark.}
Given the realizations of the optimality variables of a set of tasks for an agent $\theta$, we are interested in predicting the most probable realization of the optimality variable of a new task for this agent without observing $\theta$. We create benchmarks for this scenario by generating datasets for quiz sizes ranging from $1$ to $20$, with $5000$ examples for both training and testing. Each example is generated by randomly sampling a quiz $\mathcal{S}_{\mathrm{quiz}}$ of desired size, along with a task $s_{\mathrm{test}}$ from $\mathcal{S}_{\mathrm{init}}$, and then recording the performance of a randomly sampled agent $\theta$ on these tasks. Performance prediction techniques are evaluated by partitioning each dataset into $10$ folds, and measuring the mean prediction accuracy across the folds, along with the standard error.

\textbf{Our approach.} We perform soft-nearest neighbor matching of $s_{\mathrm{test}}$ with $\mathcal{S}_{\mathrm{quiz}}$ in the task embedding space to predict performance on $s_{\mathrm{\mathrm{test}}}$. Concretely, given the embedding function $f_{\phi}(.)$, the prediction is $\mathbf{1}_{\mathtt{c} > 0.5}$, where $\mathtt{c}$  equals $\frac{\sum_{s \in \mathcal{S}_{\mathrm{quiz}}} \mathrm{o_{s}}\,\exp(-\beta{\norm{f_{\phi}(s) - f_{\phi}(s_{\mathrm{test}})}^{2}_{2}})}{\sum_{s \in \mathcal{S}_{\mathrm{quiz}}} \exp(-\beta{\norm{f_{\phi}(s) - f_{\phi}(s_{\mathrm{test}})}^{2}_{2}})}$, $o_{s}$ is the realization of the optimality variable for task $s$, and $\beta$ is a hyperparameter. 

\textbf{Baselines.}
Besides \textit{PredModel}, we compare against different levels of oracle knowledge:
\begin{enumerate*}[label=(\roman*)]
    \item \textit{Random}: Randomly predicts the agent's performance.
    \item \textit{IgnoreTask}: Predicts the agent to succeed on $s_{\mathrm{test}}$ iff the probability that it succeeds on a random task exceeds $0.5$.
    \item \textit{IgnoreAgent}: Predicts success on $s_{\mathrm{test}}$ iff the probability that a random agent succeeds on it exceeds $0.5$.
    \item \textit{OPT}: Predicts the agent to succeed on $s_{\mathrm{\mathrm{test}}}$ iff the probability that it succeeds on $s_{\mathrm{test}}$ exceeds $0.5$.
\end{enumerate*}

\textbf{Results.}
Fig.~\ref{fig: dt_1_results} shows the prediction accuracies of various methods. Our method is competitive with \textit{OPT}, which is an upper-bound on the prediction accuracy but relies on the unrealistic assumption of full observability of both the agent and task. Additionally, our proposed task similarity criterion, which is used learn the embeddings, captures performance correlations that remain valid for new agents (see Appendix~\ref{sec:app.perf_diff_pop}).

\subsection{Task Selection}
\label{expselection}
Next, we assess the learned embeddings by using them to select tasks with desired characteristics. Specifically, we seek to answer the following research questions:  \begin{enumerate*}[label=(\roman*)]
\item Does the inner product in the learned embedding space capture task similarity according to our information-theoretic criterion?
\item Does the norm of the embedding learned by our framework induce an ordering on the tasks based on their difficulties?
\end{enumerate*} 
We begin by creating a benchmark for this scenario, and then compare our method based on task embeddings against various baselines.

\textbf{Benchmark.}
Amongst several options of tasks $\mathcal{S}_{\mathrm{options}}$, we are interested in choosing the task that best matches the desired characteristics, which we categorize into two query types: \textit{Type-$\mathit{1}$}: \textit{Select the task that is the most similar to a given reference task $s_{\mathrm{ref}}$}. The ground-truth answer to this query is $\argmax_{s \in \mathcal{S}_{\mathrm{options}}}{\mathcal{I}(\mathcal{O}_{s_{\mathrm{ref}},\Theta}; \mathcal{O}_{s,\Theta})}$. \textit{Type-$\mathit{2}$}: \textit{Select the task that is the most similar to (but harder than) a given reference task $s_{\mathrm{ref}}$}. Out of all tasks in $\mathcal{S}_{\mathrm{options}}$ that are harder than $s_{\mathrm{ref}}$, the ground-truth answer to this query is the task most similar to it. To create benchmarks for this scenario, we generate a dataset of $50$ examples. Each example consists of a randomly sampled $s_{\mathrm{ref}}$ and $10$ tasks that form $\mathcal{S}_{\mathrm{options}}$. The benchmarks also include $5$ easy tasks for reference (determined by ranking a randomly sampled pool of $500$ tasks). We evaluate task selection methods by reporting mean selection accuracy across $4$ randomly sampled datasets, along with the standard error.

\textbf{Our approach.} We use task embeddings to rank the options according to similarity and/or difficulty, based on which the selection is made. We additionally compare our method based on embeddings learned without $\mathcal{\mathcal{C}_{\mathrm{NORM}}}$ (listed as $\textit{Ours}_{\textit{woNorm}}$).

\textbf{Baselines.}
Besides \textit{PredModel}, we compare against the following baselines:
\begin{enumerate*}[label=(\roman*)]
    \item \textit{Random}: Randomly selects answers from $\mathcal{S}_{\mathrm{options}}$.
    \item \textit{StateSim}: Measures task similarity based on state representation distances. For queries of type $2$, it considers a task $s_{1}$ to be harder than $s_{2}$ iff the similarity between $s_{1}$ and the task most similar to it in the set of easy reference tasks, is less than that for $s_{2}$.
    \item \textit{TrajectorySim}: Measures task similarity using the edit distance between expert trajectories.
    \item $\mathrm{\mathit{OPT}}$: Estimates task similarity and difficulty using the entire agent population. Given the variance in the estimation process, this is the best one could do on this benchmark.
    \item $\mathrm{\widehat{\mathit{OPT}}_\mathit{50}}$: Estimates task similarity and difficulty using a randomly sampled $\mathrm{50}\%$ of the population.
\end{enumerate*}

\textbf{Results.}
Fig.~\ref{fig: dt_2_results} compares different methods on the task selection benchmark. Our method outperforms \textit{Random}, \textit{StateSim}, \textit{TrajectorySim}, and \textit{PredModel}, and is competitive with $\mathrm{\widehat{\mathit{OPT}}_\mathit{50}}$. This suggests that the inner product in the learned task embedding space successfully captures similarity between tasks. Notably, our technique significantly outperforms $\textit{Ours}_{\textit{woNorm}}$ on Type-$2$ queries, indicating that the norm of the embedding effectively orders tasks by difficulty.


\section{Concluding Discussion}
\label{sec:conclusions}
In this work, we introduced an information-theoretic framework for learning task embeddings in sequential decision-making settings. Through experiments on diverse environments, we empirically demonstrated that the inner product in the embedding space captures task similarity, and the norm of the embedding induces an ordering on the tasks based on their difficulties. A limitation of our current framework is the requirement for tasks to be goal-based, which we plan to address in future work. This could involve using the difference between the cumulative reward obtained during the rollout and the maximum achievable cumulative reward for the given task to parameterize the optimality variable. Additionally, it would be interesting to explore more principled methods for constructing the agent population that explicitly optimize for diversity. Further, empirically estimating the proposed similarity criterion by directly estimating the underlying mass functions could be sample-inefficient for some environments. Thus, a promising direction is to construct sample-efficient estimators for it. Moreover, evaluation in multi-agent settings, where the task embedding could encode the behavior of non-ego agents, is another interesting direction.
\section*{Acknowledgments}
Funded/Co-funded by the European Union (ERC, TOPS, 101039090). Views and opinions expressed are however those of the author(s) only and do not necessarily reflect those of the European Union or the European Research Council. Neither the European Union nor the granting authority can be held responsible for them.
\bibliographystyle{named}
\bibliography{main}
\clearpage
\appendix
{
    \allowdisplaybreaks
\section{Table of Contents}
\label{sec:app.toc}

In this section, we briefly describe the content provided in the paper's appendices.
\begin{itemize}[leftmargin=*]
    \item Appendix~\ref{sec:app.algo_estimate} provides the pseudocode for our algorithm to empircally estimate $\mathcal{I}$.
    \item Appendix~\ref{sec:app.env_details} provides detailed descriptions of the environments used for experimental evaluation.
    \item Appendix~\ref{sec:app.baseline_details} provides additional technical details of the \textit{PredModel} baseline.
    \item Appendix~\ref{sec:app.impl_details} provides additional implementation details for experimental evaluation.
    \item Appendix~\ref{sec:app.annotated_vis} provides annotated visualization of the task embedding spaces.
    \item Appendix~\ref{sec:app.pca_vis} provides PCA plots of the embedding spaces.
    \item Appendix~\ref{sec:app.gen_exp} provides additional results examining the embedding function's generalization ability.
    \item Appendix~\ref{sec:app.perf_diff_pop} provides additional results examining the validity of task correlations for new agents.
\end{itemize}

\section{Pseudocode for Empirical Estimation of Task Similarity}\label{sec:app.algo_estimate}
The pseudocode for the proposed algorithm to empirically estimate $\mathcal{I}$ is given in Algorithm~\ref{alg:empiricalI}. Given an MDP $\mathcal{M}$ and a prior distribution of the agent parameters $p(\theta)$, the algorithm uses $\mathtt{N}$ samples to estimate $\mathcal{I}(\mathcal{O}_{s_{i},\Theta}; \mathcal{O}_{s_{j},\Theta})$. For each sample, the algorithm randomly samples $\theta_{l} \sim p(\theta)$, and performs rollouts of $\pi_{\theta_{l}}$ from $s_{i}$ and $s_{j}$ to obtain estimates of the probability mass functions required for the computation of $\mathcal{I}$. Note that $\mathcal{H}_{b}(p)$ computes the entropy of a Bernoulli random variable $X$ s.t. $X$ takes the value $1$ with probability $p$.

\begin{algorithm}[h!]
    \caption{Empirically Estimate Task Similarity ($\mathcal{I}$)}
    \begin{algorithmic}[1]
    \small
    \Procedure{Estimate}{Task ${s}_{i}$, Task ${s}_{j}$, MDP $\mathcal{M}$, Policy $\pi$, Prior distribution of the agent parameters $p(\theta)$, Number of samples $\mathtt{N}$}
        \State $\mathtt{n\_i} \gets 0$  \Comment{\#successes on $s_{i}$}
        \State $\mathtt{n\_j} \gets 0$ \Comment{\#successes on $s_{j}$}
        \State $\mathtt{n\_i\_j\_1} \gets 0$ \Comment{\#successes on $s_{i}$ given success on $s_{j}$}
        \State $\mathtt{n\_i\_j\_0} \gets 0$ \Comment{\#successes on $s_{i}$ given failure on $s_{j}$}
        \For{$l \in \{\mathtt{1}, \dots, \mathtt{N}\}$}
                \State Sample agent parameters $\theta_{l} \sim p(\theta)$ and set it to $\pi$.
                \State Perform a rollout of $\pi_{\theta_{l}}$ from $s_{i}$ on $\mathcal{M}$.
                \State Perform a rollout of $\pi_{\theta_{l}}$ from $s_{j}$ on $\mathcal{M}$.
                \If{rollout from $s_{i}$ is a success}
                    \State $\mathtt{n\_i} \gets \mathtt{n\_i} + 1$
                    \If{rollout from $s_{j}$ is a success}
                        \State $\mathtt{n\_i\_j\_1} \gets \mathtt{n\_i\_j\_1} + 1$
                    \Else
                        \State $\mathtt{n\_i\_j\_0} \gets \mathtt{n\_i\_j\_0} + 1$
                    \EndIf
                \EndIf
                \If{rollout from $s_{j}$ is a success}
                    \State $\mathtt{n\_j} \gets \mathtt{n\_j} + 1$
                \EndIf    
        \EndFor
        \State $\hat{\mathcal{I}} \gets \mathcal{H}_{b}(\frac{\mathtt{n\_{i}}}{\mathtt{N}}) - (\frac{\mathtt{n\_{j}}}{\mathtt{N}}) \mathcal{H}_{b}(\frac{\mathtt{n\_i\_j\_1}}{\mathtt{n\_j}})  - (1 - \frac{\mathtt{n\_{j}}}{\mathtt{N}}) \mathcal{H}_{b}(\frac{\mathtt{n\_i\_j\_0}}{\mathtt{N} - \mathtt{n\_j}})$
        \State \Return $\hat{\mathcal{I}}$
        \EndProcedure  
    \end{algorithmic}
    \label{alg:empiricalI}
\end{algorithm}
\section{Environment Details}\label{sec:app.env_details}
\subsection{{\envone}}
This environment corresponds to a navigation task in a one-dimensional line segment $[0, 1]$, where the agent has to pick certain keys using appropriate $\mathtt{pickKey}$ actions (one action for each key type) and unlock the door located towards the right. A task in this environment is considered to be solved if the agent successfully unlocks the door. The environment used in our experiments is based on the work of \citet{devidze2021explicable}; however, we adapted it to have multiple keys that need to be picked.

More concretely, there are four keys, $\mathtt{KeyA}$, $\mathtt{KeyB}$, $\mathtt{KeyC}$, and $\mathtt{KeyD}$, located on the segments $[0, 0.1]$, $[0.2, 0.3]$, $[0.4, 0.5]$, $[0.6, 0.7]$, respectively. A door is located on the segment $[0.9, 1]$. The door could be of the following $4$ types: Type $1$ ($\mathtt{00}$), Type $2$ ($\mathtt{01}$), Type $3$ ($\mathtt{10}$), or Type $4$ ($\mathtt{11}$). Doors of Type $1$ require $\mathtt{KeyA}$ and $\mathtt{KeyB}$, doors of Type $2$ require $\mathtt{KeyA}$ and $\mathtt{KeyC}$, doors of Type $3$ require $\mathtt{KeyB}$ and $\mathtt{KeyD}$, and doors of Type $4$ require $\mathtt{KeyC}$ and $\mathtt{KeyD}$. 

The set of initial states, $\mathcal{S}_{\mathrm{init}}$, is the same as the set of states $\mathcal{S}$. Each state $s \in \mathcal{S}$ corresponds to a $7$-tuple ($\mathtt{Location}$, $\mathtt{KeyStatusA}$, $\mathtt{KeyStatusB}$, $\mathtt{KeyStatusC}$, $\mathtt{KeyStatusD}$, $\mathtt{doorBit1}$, $\mathtt{doorBit2}$). Here, $\mathtt{Location}$ denotes the agent's location on the line segment, and $\mathtt{KeyStatusA}$, $\mathtt{KeyStatusB}$, $\mathtt{KeyStatusC}$, $\mathtt{KeyStatusD}$ are flags for whether the agent has picked up the corresponding key. Task variability in this environment comes from the agent's initial position, the keys that it possesses initially, and the door type (with each type requiring a unique set of keys).

The action space is $\mathcal{A} = \{$$\mathtt{moveLeft}$, $\mathtt{moveRight}$, $\mathtt{pickKeyA}$, $\mathtt{pickKeyB}$, $\mathtt{pickKeyC}$, $\mathtt{pickKeyD}$, $\mathtt{finish}$$\}$. $\mathtt{moveLeft}$ and $\mathtt{moveRight}$ move the agent across the environment with step size $0.075 + \epsilon$, where $\epsilon~\sim~\mathit{U}(-0.01, 0.01)$. If $\mathtt{pickKeyA}$ is executed at a location that lies on the segment containing $\mathtt{KeyA}$, $\mathtt{KeyStatusA}$ becomes $\mathtt{True}$, else the environment crashes. Likewise for $\mathtt{pickKeyB}$, $\mathtt{pickKeyC}$, and $\mathtt{pickKeyD}$. The agent gets a reward of $1$ on executing $\mathtt{finish}$ if it is at a location that lies on the segment containing the door and possesses the required keys; $\mathtt{finish}$ results in a crash otherwise. The horizon length is $40$ and $\gamma$ is $0.999$.

\subsection{{\envtwo}}
This environment is a variation of the classic control task from OpenAI gym \citep{gym}, and also takes inspiration from \citep{Sodhani2021MTEnv} in which the forces applied by each action could be negative as well. It consists of a pole (mass: $0.1$ $\mathrm{kg}$, length:  $1$ $\mathrm{m}$) attached by an unactuated joint to a cart (mass: $1$ $\mathrm{kg}$), which moves along a track. The agent controls the cart in the presence of gravity (g: $9.8$ $\mathrm{m/s^{2}}$) by applying forces to the the left (action $0$) or to the right (action $1$) of the cart. A task is considered to be solved if the agent keeps the pole upright for $200$ timesteps.

Each state $s \in \mathcal{S}$ corresponds to a tuple $(\mathtt{x}$, $\mathtt{v}$, $\theta$, $\omega$, $\mathtt{F}$, $\mathtt{TaskType}$, $\mathtt{NumSteps})$. Here, $\mathtt{x}$ denotes the position of the cart, $\mathtt{v}$ denotes the velocity of the cart, $\theta$ denotes the angle that the pole makes with the vertical, $\omega$ denotes the angular velocity of the pole, $\mathtt{F} \in {[-15\,\mathrm{N}, -5\,\mathrm{N}] \cup [5\,\mathrm{N}, 15\,\mathrm{N}]}$ denotes the force applied on the cart by each action, $\mathtt{TaskType} \in \{0, 1\}$ denotes the type of the task, and $\mathtt{NumSteps}$ denotes the number of steps passed since the beginning of the episode.

Task variability in this environment comes from the force $\mathtt{F}$ applied on the cart by each action, and the $\mathtt{TaskType} \in \{0, 1\}$. Tasks of Type $0$ involve ``Pulling'' with action $0$ pulling the cart from the left and action $1$ pulling the cart from the right, while tasks of Type $1$ involve ``Pushing''. At any timestep, if $\theta \notin [-12^{\circ}, 12^{\circ}]$, the pole is not upright and consequently, the environment crashes. The horizon length is $200$ and $\gamma$ is $1$.

\subsection{{\envfive}}
This environment was introduced by \citet{sprl}. We provide details here for completeness. 

The agent applies forces to control a point mass inside a square space $[-4, 4] \times [-4, 4]$ surrounded by walls. The space exhibits friction, with the coefficient of kinetic friction $\mathtt{\mu_{k}} \in [0, 4]$. Additionally, there is a gate of width $\mathtt{w_{g}} \in [-4, 4]$ at position $\mathtt{p_{g}} \in [0.5, 8]$, effectively spanning the segment $[\mathtt{p_{g}} - 0.5 \mathtt{w_{g}}, \mathtt{p_{g}} + 0.5 \mathtt{w_{g}}]$. The agent always starts off from the fixed initial position $[0, 3]$. A task in this environment is considered to be solved if the point mass reaches the fixed goal position $[0, -3]$, which requires crossing the gate.

Each state $s \in \mathcal{S}$ corresponds to a tuple $(\mathtt{x}, \mathtt{v_x}, \mathtt{y}, \mathtt{v_y})$. Here, $[x, y]$ denotes the position of the point mass, while $[\mathtt{v_x}, \mathtt{v_y}]$ denotes the velocity. The actions are $[\mathtt{F}_x, \mathtt{F}_y] \in {[-10, 10] \times [-10, 10]}$, where $\mathtt{F}_x$ and $\mathtt{F}_y$ correspond to forces applied along the $x$ and the $y$ axis, respectively. Task variability in this environment comes from the width $\mathtt{w_{g}}$ and the position $\mathtt{p_{g}}$ of the gate, along with the coefficient of kinetic friction $\mathtt{\mu_{k}}$ of the space. At any timestep, if the point mass crashes into the wall, the environment crashes. The horizon length is $100$ and $\gamma$ is $0.99$.

\subsection{{\envfour}}
This is the Karel program synthesis environment from \citep{iclr18}. Karel is an educational programming language widely used in introductory CS courses. The environment consists of an avatar (characterized by its position and orientation) inside an $18 \times 18$ grid in which each cell could contain up to $10$ markers, correspond to a wall, or be empty. The avatar can move inside the grid and modify it by picking or placing markers. The objective of each task is to synthesize the program ${\pi}^{\ast}$ (which is a controller for the avatar) in the Karel domain-specific language (DSL) given $5$ input-output examples for it in the form of $\mathtt{Pre\text{-}Grid}$ and its corresponding $\mathtt{Post\text{-}Grid}$. A task is considered to be solved if the synthesized program $\pi$ generalizes to a held-out test example for ${\pi}^{\ast}$. 

 The Karel DSL is shown in Figure~\ref{fig:app.karel_dsl}. Task variability in this environment comes from the set of input-output examples. The state in this environment comprises of the program specification (i.e., the input-output examples) and the partial program synthesized so far. The tokens of the DSL form the action space. The horizon length is $24$ and $\gamma$ is $1$. We use a set of $73688$ tasks (with the number of tokens in ${\pi}^{\ast}$ ranging between 10 and 14) sampled from the dataset used in \citep{iclr18} (accessible at \url{https://msr-redmond.github.io/karel-dataset/}). 

\begin{figure}[h]
\centering
\includegraphics[width=\columnwidth]{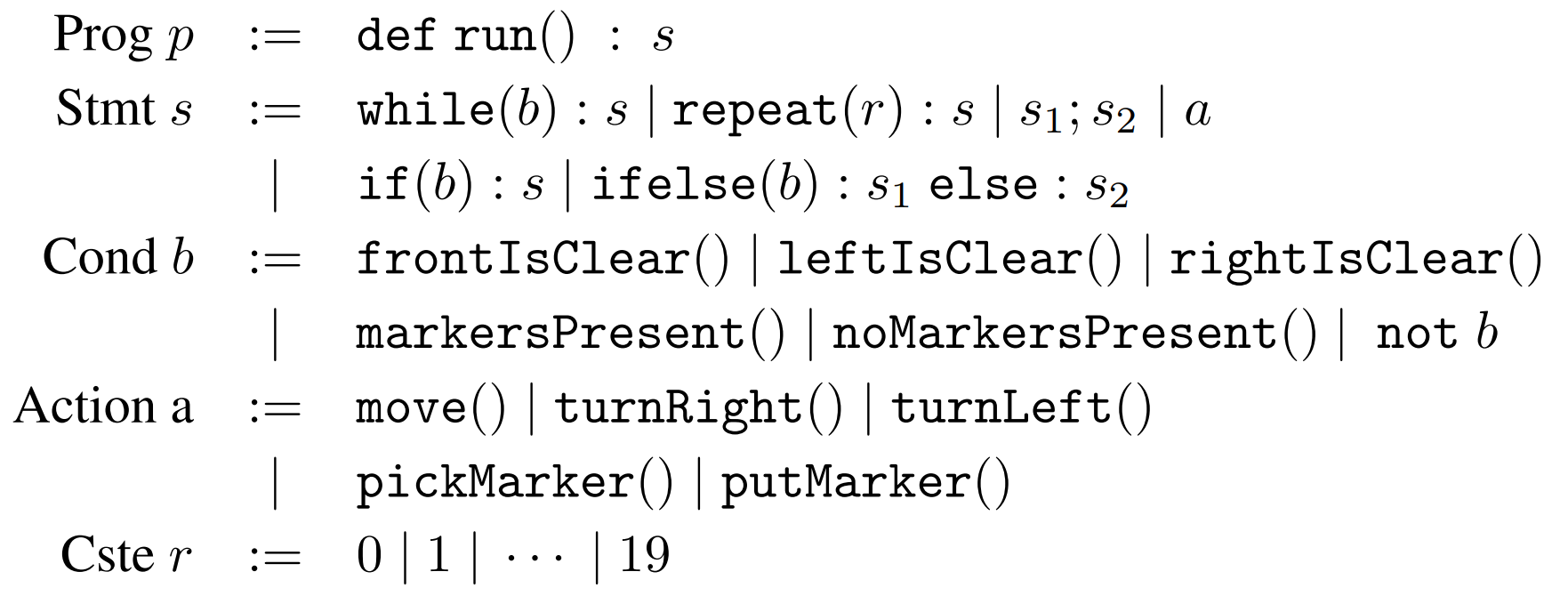}
\caption{The Karel DSL \protect\citep{iclr18}.}
\label{fig:app.karel_dsl}
\end{figure}

\begin{figure*}[!ht]
\centering

   \begin{subfigure}[t]{\textwidth}
 \centering
    \includegraphics[width=0.7\textwidth, trim={-1cm 0 0.05cm 0}, clip]{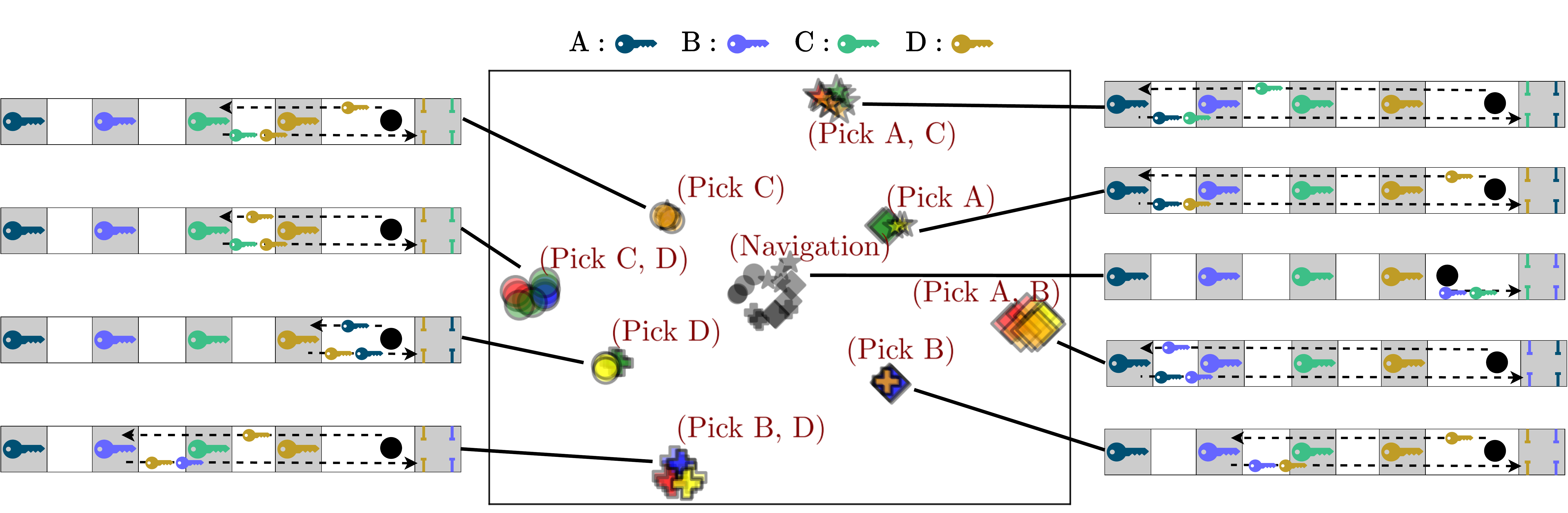}
    \caption{\envone}
    \label{fig: app.annotated_vis_MultiKeyNav}
  \end{subfigure}
 \begin{subfigure}[t]{0.465\textwidth}
  \centering
    \includegraphics[width=\textwidth, trim={0 0 5cm 0}, clip]{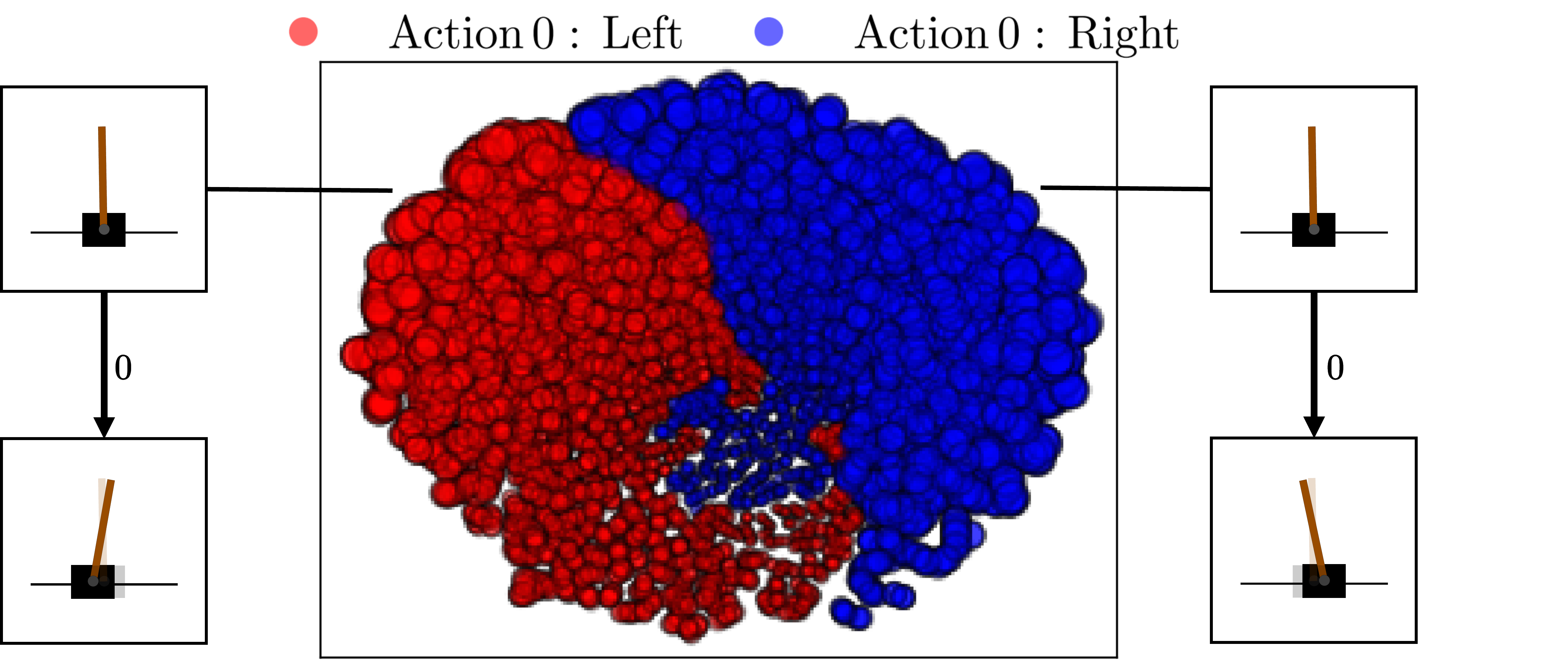}
    \caption{\envtwo}
    \label{fig: app.annotated_vis_CartPoleVar}
  \end{subfigure}
  \begin{subfigure}[t]{0.47\textwidth}
  \centering
    \includegraphics[width=\textwidth, trim={0 0 0.3cm 0}, clip]{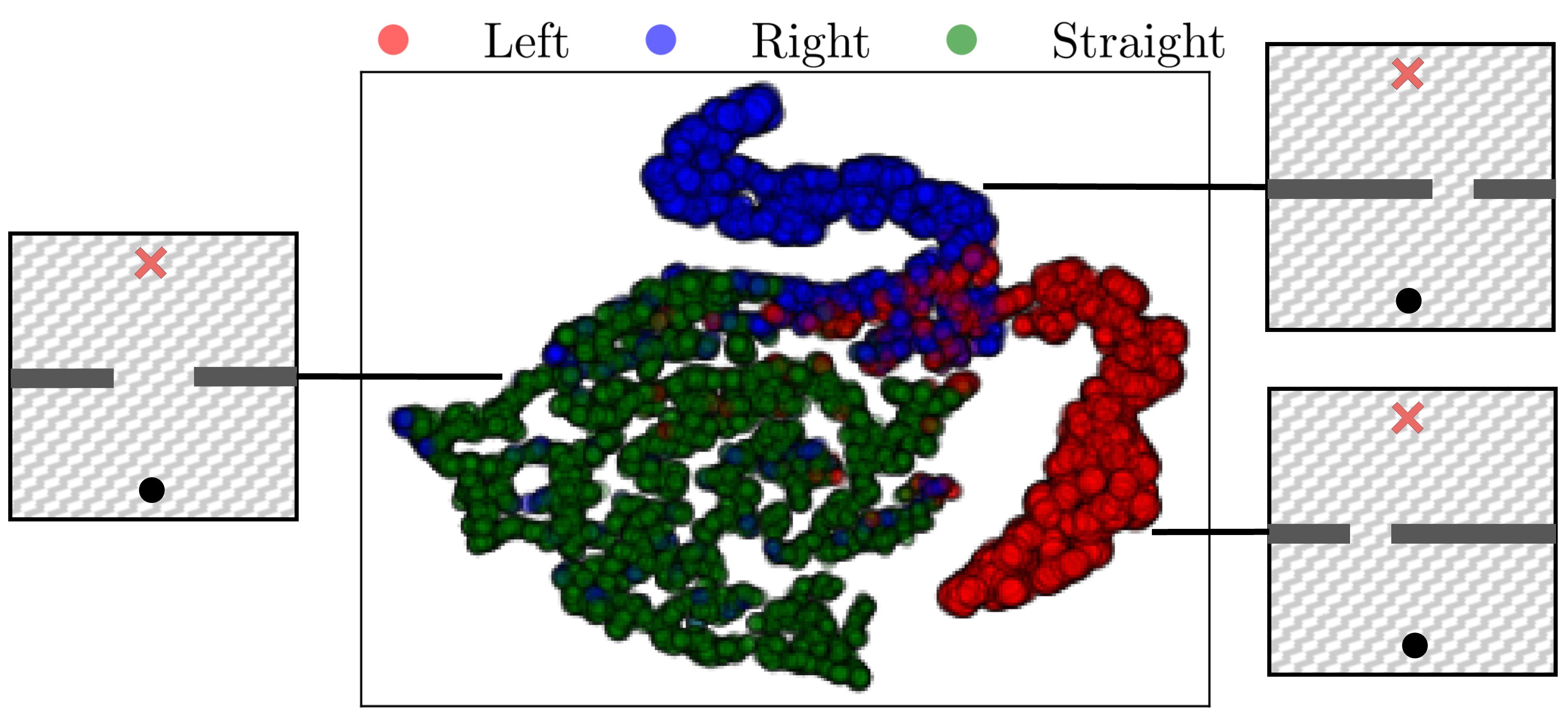}
    \caption{\envfive}
    \label{fig: app.annotated_vis_PointMass}
  \end{subfigure}
  \begin{subfigure}[t]{\textwidth}
  \centering
    \includegraphics[width=0.93\textwidth, trim={0 0 3.5cm 0}, clip]{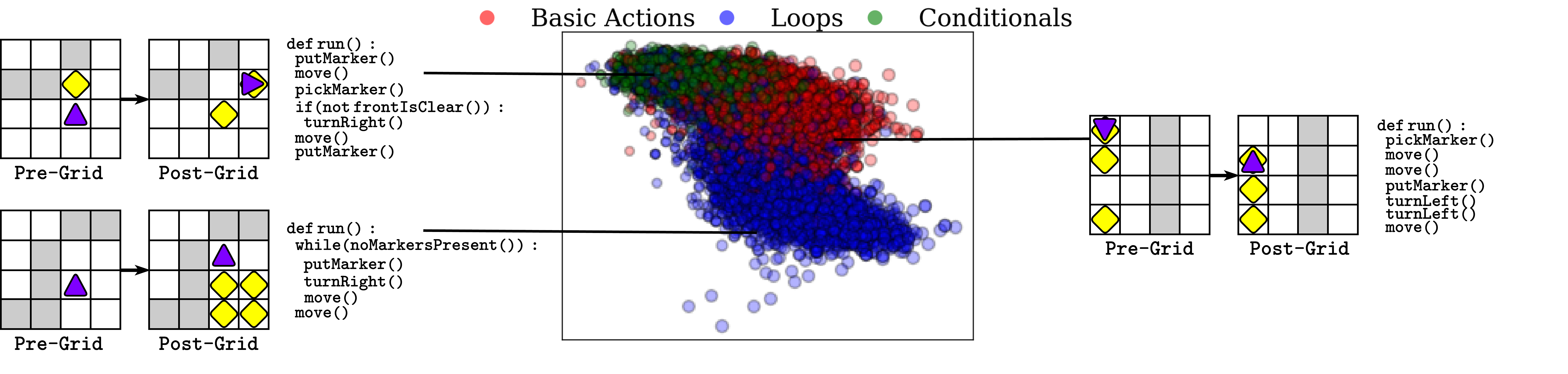}
    \caption{\envfour}
    \label{fig: app.annotated_vis_Karel}
  \end{subfigure}  
  \begin{subfigure}[t]{\textwidth}
  \centering
    \includegraphics[width=0.675\textwidth, trim={-1.95cm 0 7cm 0}, clip]{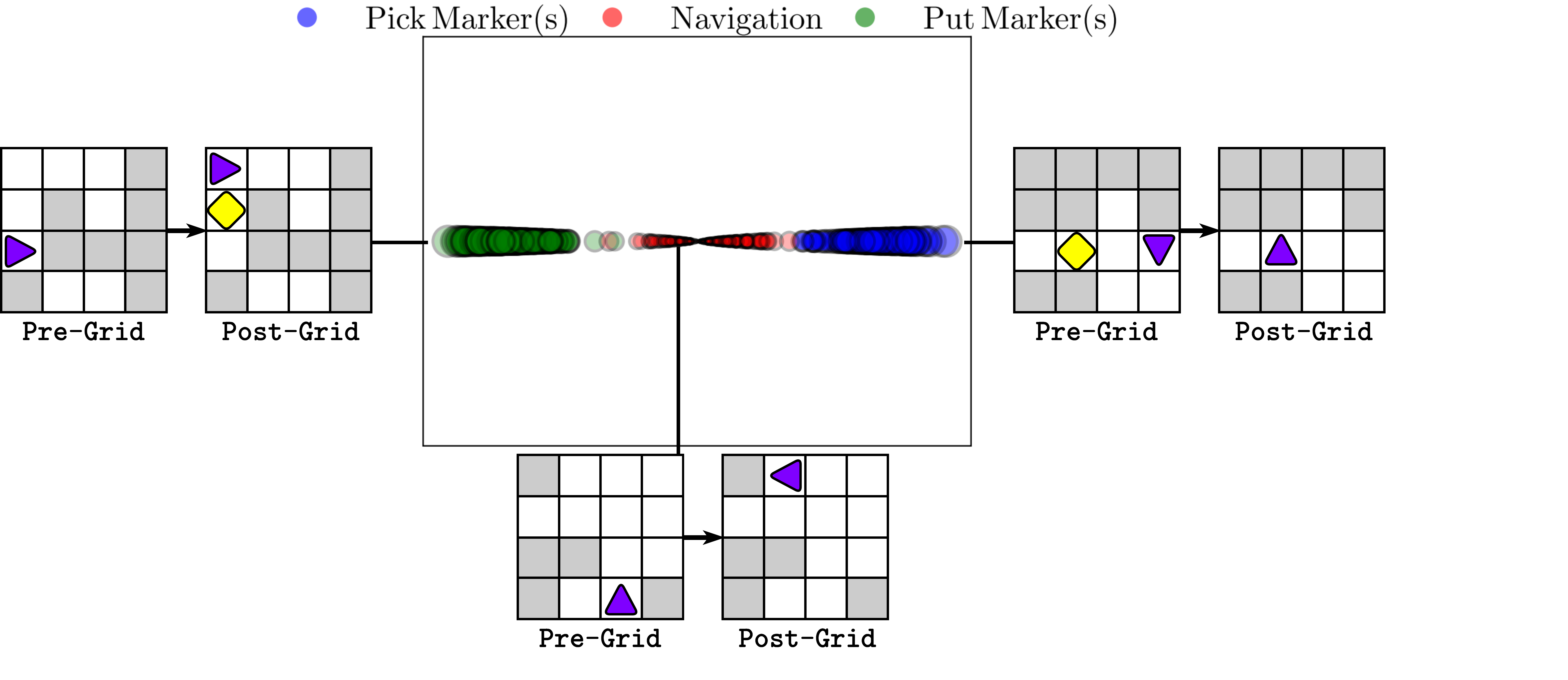}
    \caption{\envthree}
    \label{fig: app.annotated_vis_BasicKarel}
  \end{subfigure}
  
\caption{Annotated visualization of the task embedding spaces with example tasks for each cluster.}
  \label{fig: app.annotated_vis}
  \vspace{-10pt}
\end{figure*}

\subsection{{\envthree}}
This environment, introduced by \citet{tzannetos2023proximal}, is a variant of {\envfour} that excludes control flow constructs such as loops and conditionals, and only includes basic actions. We provide the details here for completeness.

The environment consists of an agent inside a $4 \times 4$ grid. Each cell in the grid could contain a marker, correspond to a wall, or be empty. The objective of each task is to generate a sequence of actions that transforms a pre-grid to a post-grid. The {\envthree} dataset has $24000$ training tasks and $2400$ validation tasks. The set of initial states, $ \mathcal{S}_{\mathrm{init}}$, is the same as the training set of tasks provided in the {\envthree} dataset. Each state $s \in \mathcal{S}$ corresponds to a tuple $(\mathtt{Curr\text{-}Grid, Post\text{-}Grid})$, where $\mathtt{Curr\text{-}Grid}$ and $\mathtt{Post\text{-}Grid}$ correspond to the bitmap representation of the current-grid and the post-grid, respectively. 

The action space is $\mathcal{A} = \{$$\mathtt{move}$, $\mathtt{turnLeft}$, $\mathtt{turnRight}$, $\mathtt{pickMarker}$, $\mathtt{putMarker}$, $\mathtt{finish}$$\}$. $\mathtt{move}$ moves the agent in the facing direction. $\mathtt{turnLeft}$ and $\mathtt{turnRight}$ turn the agent left and right, respectively. The agent can pick and put a marker using $\mathtt{pickMarker}$ and $\mathtt{putMarker}$, respectively. The agent gets a reward of $1$ on executing $\mathtt{finish}$ if $\mathtt{Curr\text{-}Grid}$ matches $\mathtt{Post\text{-}Grid}$ (i.e., it has successfully transformed the pre-grid to the post-grid); $\mathtt{finish}$ results in a crash otherwise. The horizon length is $20$ and $\gamma$ is $0.999$.
\section{Technical Details of the PredModel Baseline}\label{sec:app.baseline_details}

\textbf{Method.} We begin by constructing a dataset $\mathcal{D}$ comprising transitions $(s_0, \bar{s}_t, a_t, r_{t+1}, \bar{s}_{t+1})$, obtained from $\mathrm{N_R}$ rollouts (with each rollout comprising multiple transitions) of the expert multi-task policy in the MDP from randomly sampled tasks, where $s_0$ represents a task, and $\bar{s}$ denotes state $s$ with context variables removed. The task embedding $z$ is inferred through variational inference.

More concretely, we train an inference network $q_\phi(z \mid s_0)$, parameterized by $\phi$, which is modeled as a diagonal Gaussian approximate posterior over z. This network is trained to reconstruct the MDP through predictive models of reward $f_{\theta_r}^{(r)}$ and dynamics $f_{\theta_s}^{(s)}$, both parameterized by $\theta_r$ and $\theta_s$, respectively. Essentially, we solve the following optimization problem: 

\begin{equation}
\begin{aligned}
    &\min_{\phi, \theta_r, \theta_s}\Biggl[\mathop{\mathbb{E}}_{(s_0, \bar{s}_t, a_t, r_{t+1}, \bar{s}_{t+1}) \sim \mathcal{D}}\biggl[\beta D_{\text{KL}}(q_\phi(z|s_0) \| p(z))\\
    &+ \mathop{\mathbb{E}}_{z \sim q_\phi(z \mid s_0)}\bigl[\alpha_r \|f_{\theta_r}^{(r)}(\bar{s}_t, a_t, z) - r_{t+1}\|_2^{2}\\
    &+ \alpha_s \|f_{\theta_s}^{(s)}(\bar{s}_t, a_t, z) - s_{t+1}\|_2^{2}\bigr]\biggr]\Biggr], \notag
\end{aligned}
\end{equation}
\normalsize
where $p(z)$ is a standard normal prior over $z$, and $\alpha_r$, $\alpha_s$, and $\beta$ are hyperparameters.

\textbf{Implementation details.} For each environment, $\mathrm{N_R}$ is set to $10000$, and the inference network, as well as the predictive models, are implemented as feedforward neural networks with $2$ hidden layers ($128$ neurons in each layer) and ReLU activations. The predictive models share weights, except for the final layer. We set $\alpha_r$, $\alpha_s$, and $\beta$, to $1$, $1$, and $0.01$, respectively. The networks are jointly trained for $500$ epochs using the Adam optimizer with $0.001$ learning rate and a batch size of $512$. The embedding dimensionalities are set to $6$, $3$, $3$, and $8$, for {\envone}, {\envtwo},  {\envfive}, and {\envthree}, respectively. 
\vspace{40pt}
\begin{figure*}[!ht]
\centering

   \begin{subfigure}{0.325\textwidth}
 \centering
    \includegraphics[width=\textwidth,]{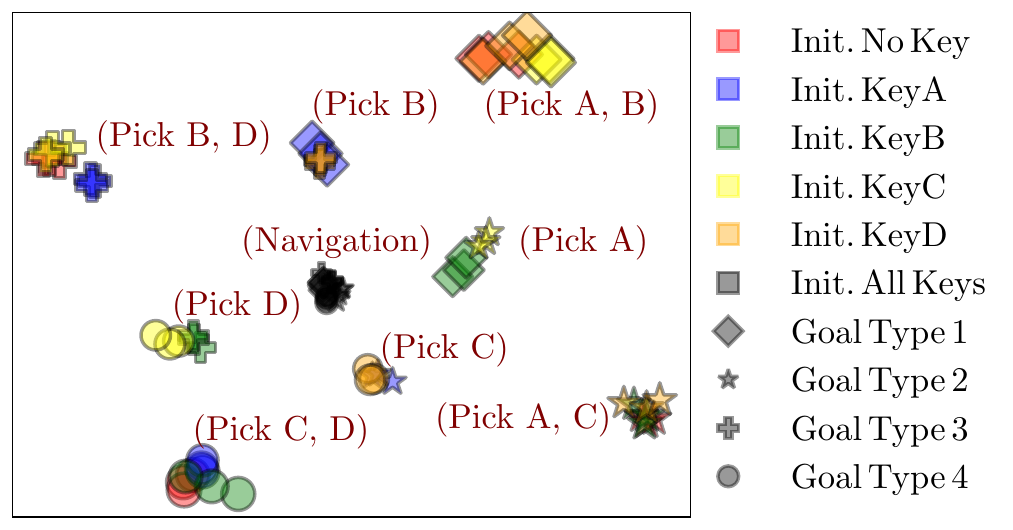}
    \caption{\envone}
    \label{fig: pca_vis_MultiKeyNav_default}
  \end{subfigure}
 \begin{subfigure}{0.25\textwidth}
  \centering
    \includegraphics[width=\textwidth]{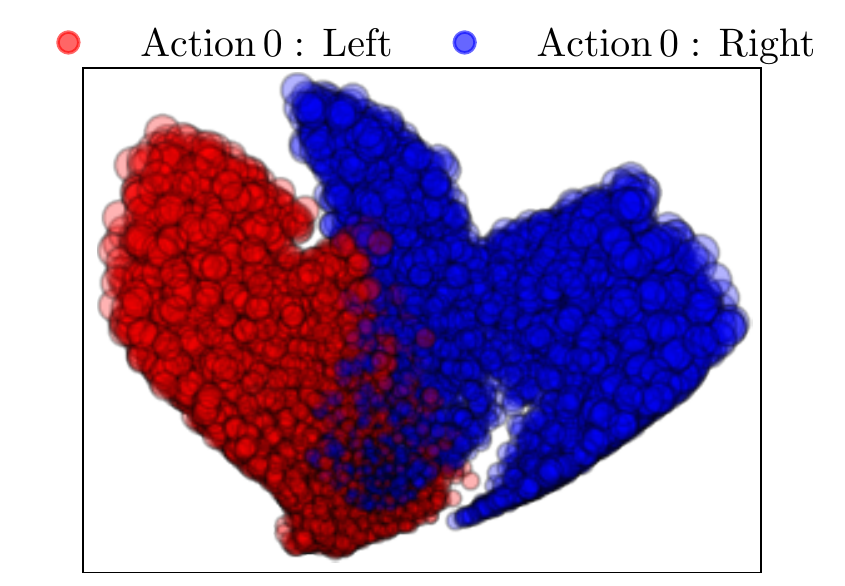}
    \caption{\envtwo}
    \label{fig: pca_vis_CartPoleVar}
  \end{subfigure} \\
\vspace{10pt}
  \begin{subfigure}{0.25\textwidth}
    \includegraphics[width=\textwidth]{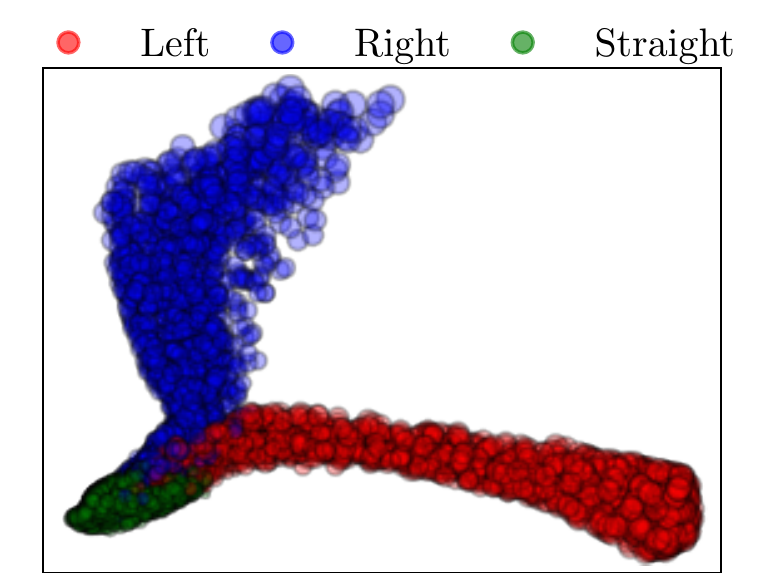}
    \caption{\envfive}
    \label{fig: pca_vis_PointMass}
  \end{subfigure}
  \begin{subfigure}{0.348\textwidth}
    \includegraphics[width=\textwidth]{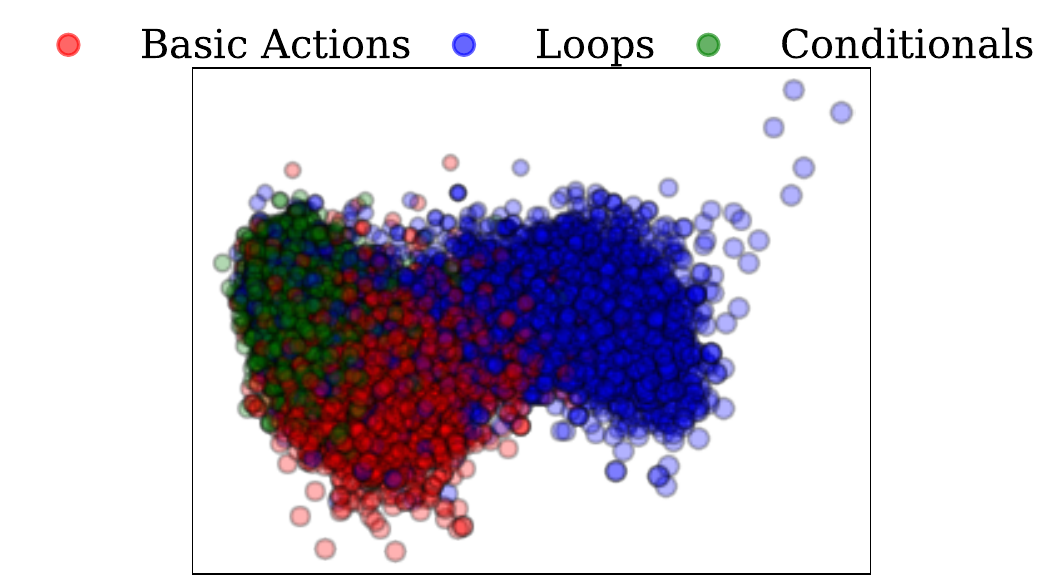}
    \caption{\envfour}
    \label{fig: pca_vis_Karel}
  \end{subfigure}  
  \begin{subfigure}{0.36\textwidth}
    \includegraphics[width=\textwidth]{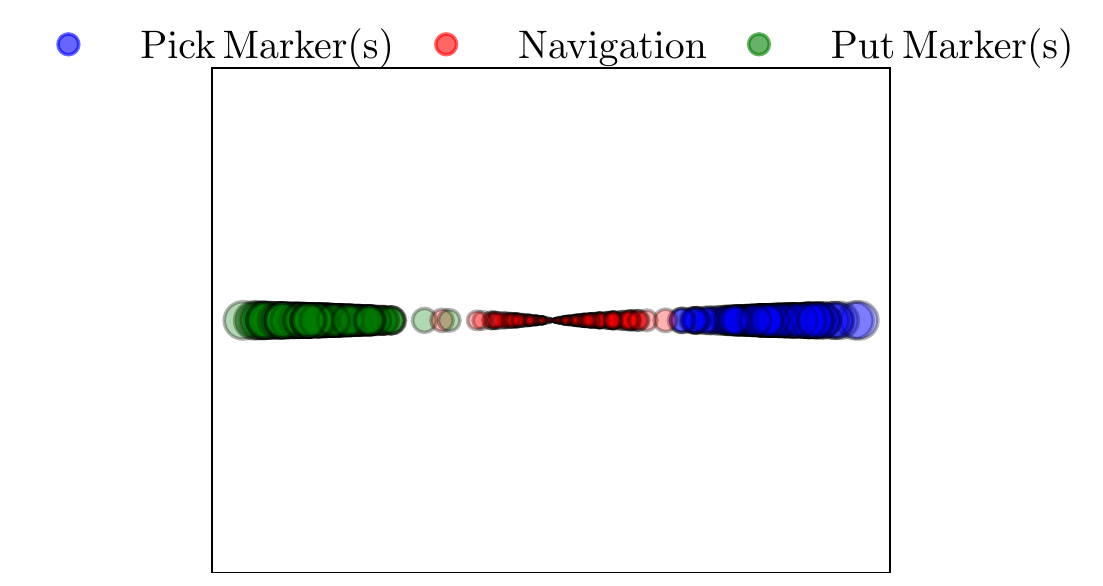}
    \caption{\envthree}
    \label{fig: pca_vis_BasicKarel}
  \end{subfigure}
  
\caption{PCA projections of the task embedding spaces learned through our framework. Each point represents a task, and the size of the points is proportional to the norm of the embeddings.}
  \label{fig: pca_vis}
  \vspace{-10pt}
\end{figure*}

\vspace{-50pt}
\section{Implementation Details}\label{sec:app.impl_details}

\textbf{Compute resources.} All the experiments were conducted on a cluster of machines with Intel Xeon Gold 6134M CPUs (clocked at $3.20$ Ghz) and Nvidia Tesla V100 GPUs ($32$ GB VRAM configuration). We would like to highlight that learning the task embedding function is a one-time process, which could be compute intensive. However, once the training is completed, computing the embeddings or utilizing them to measure task similarity is a one-shot operation.

\textbf{Training process.} For all the environments, we set $\delta_{\mathrm{snap}}$ to $0.01$ ($0.1$ for {\envfour}) in our experiments. The average performance on $\mathcal{S}_{\mathrm{snap}}$ is determined by performing $10$ rollouts on each task in the set. We implement each embedding network using a succession of several fully connected layers with $\mathtt{ReLU}$ activations, trained for $300$ epochs ($500$ epochs for {\envtwo} and $10$ epochs for {\envfour}) using the Adam optimizer with $\mathrm{1e-3}$ ($\mathrm{1e-4}$ for {\envfour}) learning rate, and batch size $128$ ($512$ for {\envfour}). We sample $5000$ ($80000$ for {\envfour} and $10000$ for {\envthree}) constraints from $\mathcal{C}_{\mathrm{MI}}$ and $\mathcal{C}_{\mathrm{NORM}}$ each to train the network, the $\hat{\mathcal{I}}$ values of which are approximated using $100$ samples from each agent in the population. $\mathrm{p\_success}(.)$ is approximated using $10$ samples from each agent in the population. The hyperparameters $\beta$ and $\lambda$ are set to $1000$ and $0.4$, respectively. We use a validation set and a test set, consisting of $1000$ ($16000$ for {\envfour} and $2000$ for {\envthree}) constraints from $\mathcal{C}_{\mathrm{MI}}$ and $\mathcal{C}_{\mathrm{NORM}}$ each, for early stopping and to determine the final model parameters. We vary the embedding dimensionality from $1$ to $10$ and choose the one after which the test loss does not decrease much. Below we provide environment-specific details:
\begin{itemize}
    \item \textbf{{\envone}}: $\mathcal{S}_{\mathrm{snap}}$ includes all combinations of the locations $\{0.05, 0.45, 0.85\}$, key statuses, and door types. We combine the subpopulations obtained by masking no action, masking each $\mathtt{pickKey}$ action individually, and masking all $\mathtt{pickKey}$ actions, to obtain an agent population of size $100$. The embedding network has two hidden layers ($32$ neurons in each layer). The embedding dimensionality is $5$ without $\mathcal{C}_{\mathrm{NORM}}$, and $6$ with $\mathcal{C}_{\mathrm{NORM}}$.
    \item \textbf{{\envtwo}}: $\mathcal{S}_{\mathrm{snap}}$ consists of $1000$ tasks sampled from $ \mathcal{S}_{\mathrm{init}}$. We combine the subpopulations obtained by using all tasks in $ \mathcal{S}_{\mathrm{init}}$, tasks in $ \mathcal{S}_{\mathrm{init}}$ with $+\text{ve }\mathtt{F}$ and Type $0$, tasks in $ \mathcal{S}_{\mathrm{init}}$ with $+\text{ve }\mathtt{F}$ and Type $1$, tasks in $ \mathcal{S}_{\mathrm{init}}$ with $-\text{ve }\mathtt{F}$ and Type $0$, and tasks in $ \mathcal{S}_{\mathrm{init}}$ with $-\text{ve }\mathtt{F}$ and Type $1$, to obtain an agent population of size $95$. The embedding network has two hidden layers ($64$ neurons in the first layer and $32$ neurons in the second). The embedding dimensionality is $2$ without $\mathcal{C}_{\mathrm{NORM}}$, and $3$ with $\mathcal{C}_{\mathrm{NORM}}$.
    \item \textbf{{\envfive}}: $\mathcal{S}_{\mathrm{snap}}$ consists of $100$ tasks sampled from $ \mathcal{S}_{\mathrm{init}}$. We combine subpopulations obtained by using all the tasks in $ \mathcal{S}_{\mathrm{init}}$, tasks in $\mathcal{S}_{\mathrm{init}}$ that satisfy the condition $\mathtt{p_{g}} + 0.5 \mathtt{w_{g}} < 0$, and tasks in $\mathcal{S}_{\mathrm{init}}$ that satisfy the condition $\mathtt{p_{g}} + 0.5 \mathtt{w_{g}} \geq 0$, to obtain an agent population of size about $25$. The embedding network has two hidden layers ($32$ neurons in each layer). The embedding dimensionality is $3$ with $\mathcal{C}_{\mathrm{NORM}}$.
    \item \textbf{{\envfour}}: $\mathcal{S}_{\mathrm{snap}}$ consists of $14681$ tasks sampled from the Karel dataset. We combine subpopulations obtained by using all the tasks in $ \mathcal{S}_{\mathrm{init}}$, tasks in $ \mathcal{S}_{\mathrm{init}}$ that do not require synthesizing token for loops, tasks in $ \mathcal{S}_{\mathrm{init}}$ that do not require synthesizing tokens for conditionals, and tasks in $ \mathcal{S}_{\mathrm{init}}$ that do not require synthesizing tokens for both loops and conditionals, to obtain an agent population of size about $135$. We use the official codebase of \citep{iclr18} to train the agents. The embedding network consists of an input-output encoder (which is the same as that in \citep{iclr18}) followed by a feedforward network with a single hidden layer ($256$ neurons). The embedding dimensionality is $2$ with $\mathcal{C}_{\mathrm{NORM}}$.
    \item \textbf{\envthree}: $\mathcal{S}_{\mathrm{snap}}$ consists of all the $2400$ validation tasks. We combine the subpopulations obtained by masking no action, masking $\mathtt{pickMarker}$, masking $\mathtt{putMarker}$, and masking both $\mathtt{pickMarker}$ and $\mathtt{putMarker}$, to obtain an agent population of size about $55$. The embedding network has two hidden layers ($32$ neurons in each layer). The embedding dimensionality is $1$ with $\mathcal{C}_{\mathrm{NORM}}$. 
\end{itemize}

\textbf{Performance prediction benchmark.} We use $500$ samples to estimate the \textit{IgnoreTask} baseline, $10$ samples from each agent in the population to estimate the \textit{IgnoreAgent} baseline, and $10$ samples to estimate the \textit{OPT} baseline. $\beta$ is tuned using the training examples.

\textbf{Task selection benchmark.} We use $100$ samples from each agent in the population to estimate task similarity, and $10$ samples from each agent to estimate task difficulty.
\section{Annotated Visualization of Embedding Spaces}\label{sec:app.annotated_vis}
In Fig.~\ref{fig: app.annotated_vis}, we visualize the learned embedding spaces annotated with an example task for each cluster.
\begin{figure}[t]
\centering

   \begin{subfigure}[t]{0.7\columnwidth}
 \centering
    \includegraphics[width=\textwidth]{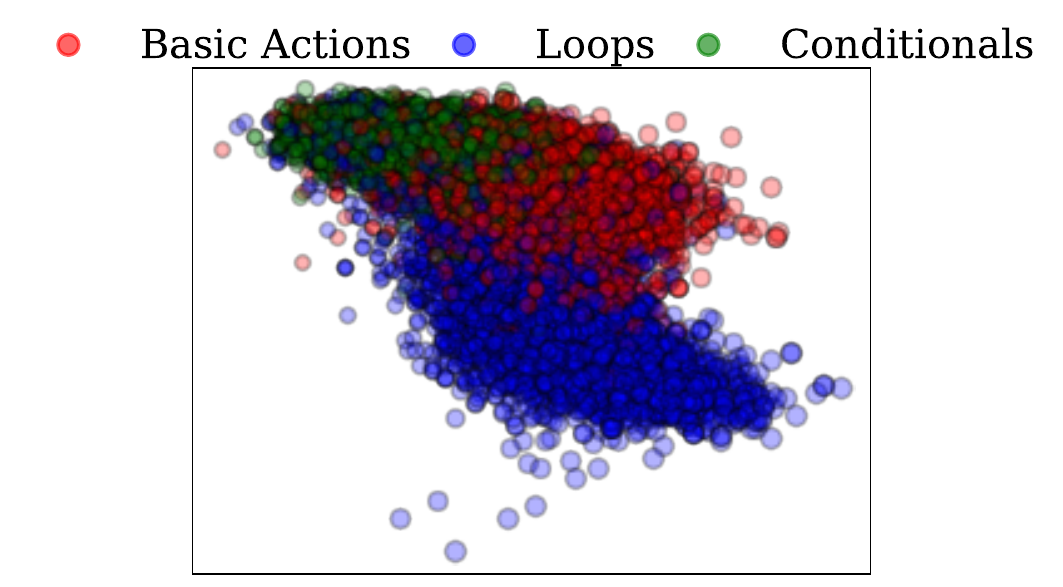}
    \caption{Training Set}
    \label{fig: gen_exp_train}
  \end{subfigure}
 \begin{subfigure}[t]{0.7\columnwidth}
  \centering
    \includegraphics[width=\textwidth]{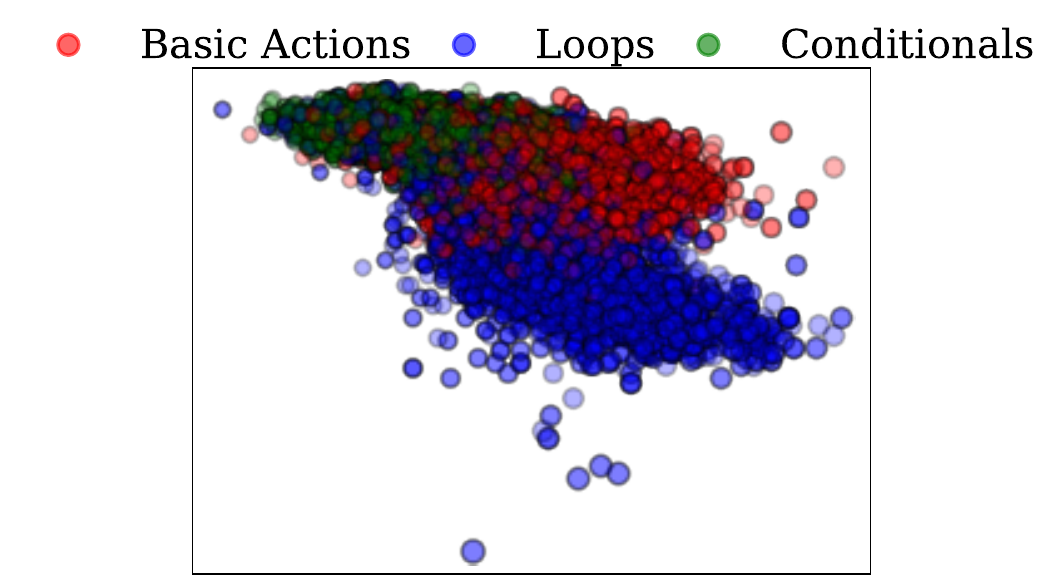}
    \caption{Validation Set}
    \label{fig: gen_exp_val}
  \end{subfigure}
  
\caption{Task embedding spaces obtained using the training and validation sets of tasks for the {\envfour} environment. This visualization shows that an embedding function learned using a training set of tasks generalizes to a validation set of tasks.}
  \label{fig: app.gen_exp_vis}
  \vspace{-10pt}
\end{figure}

\vspace{-5pt}
\section{PCA Plots}\label{sec:app.pca_vis}
Fig.~\ref{fig: pca_vis} visualizes the learned task embedding spaces through 2D PCA projections.
\vspace{-5pt}
\section{Generalization Experiment}\label{sec:app.gen_exp}
We present a generalization experiment in which we assess if the embedding function (learned using a training set of tasks) produces a consistent embedding space for a validation set of tasks. We conduct this evaluation on the {\envfour} environment by partitioning the tasks into training and validation sets of size $59007$ and $14681$, respectively. Fig.~\ref{fig: app.gen_exp_vis} visualizes the task embedding spaces for these sets and shows the generalization ability of the learned task embedding function.
\section{Performance Prediction -- New Agents}\label{sec:app.perf_diff_pop}
\begin{figure}[t]
\centering
\includegraphics[width=0.8\columnwidth]{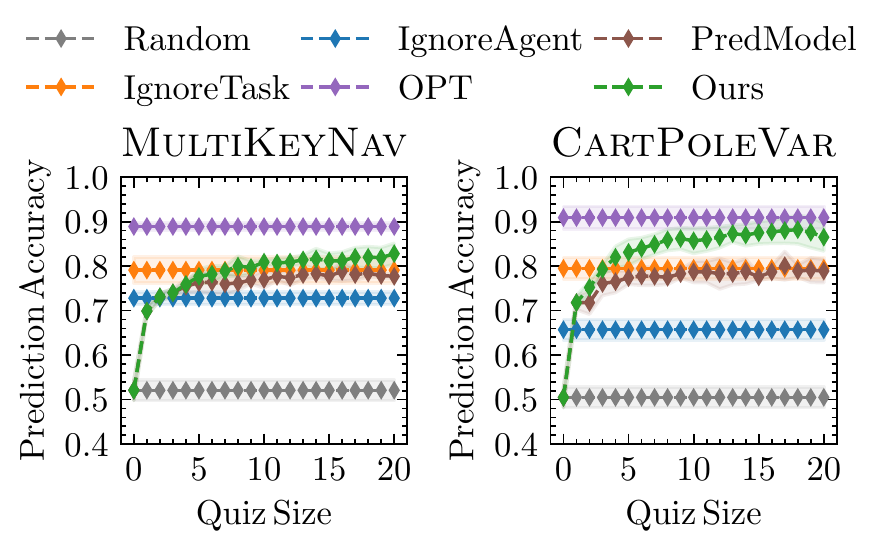}
\caption{Performance prediction results for new agents using task embeddings. Our technique is competitive with the \textit{OPT} baseline, which is the best one could do on this benchmark.}
\label{fig: app.dt_1_diff_pop}
\vspace{-10pt}
\end{figure}

To assess if the correlations captured by our similarity criterion remain valid for new agents, we evaluate the learned task embedding network in the performance prediction application scenario for new agents. For {\envone}, the new population of agents is created using biased task distributions instead of action masking. More concretely, we combine the subpopulations obtained by using all tasks, tasks with doors of Type $1$, tasks with doors of Type $2$, tasks with doors of Type $3$, and tasks with doors of Type $4$. For {\envtwo}, we use the Proximal Policy Optimization (PPO) algorithm \citep{Schulman2017ProximalPO} instead of behavioral cloning to create the new agent population.

Fig.~\ref{fig: app.dt_1_diff_pop} shows the prediction accuracies of various techniques. Our method is competitive with \textit{OPT}, which demonstrates the efficacy of task embeddings in predicting the performance of new agents.
}

\end{document}